%% file: main.tex
\documentclass{article} 
\usepackage[final]{acl}
\usepackage{times}
\usepackage{latexsym}
\input{math_commands.tex}

\usepackage{amsmath}
\usepackage{amssymb}
\usepackage{amsfonts}
\usepackage{amsthm}

\usepackage{booktabs}
\usepackage{array}
\usepackage{tabularx}
\usepackage{multirow}
\usepackage{multicol}
\usepackage{makecell}      
\usepackage{threeparttable}
\usepackage{colortbl}
\usepackage{arydshln}      
\usepackage{bigstrut}
\usepackage{xcolor}

\usepackage{graphicx}
\usepackage{adjustbox}
\usepackage{caption}
\usepackage{subcaption}
\usepackage{sidecap}
\usepackage{placeins}      

\usepackage{tikz}
\usetikzlibrary{tikzmark}
\usepackage{pgffor}

\usepackage{xcolor}
\usepackage{xspace}
\usepackage{enumitem}
\usepackage{pifont}
\usepackage{verbatim}      
\usepackage{comment}
\usepackage{boxedminipage}
\usepackage{eqparbox}
\usepackage{listings}
\usepackage[linewidth=1pt]{mdframed}
\usepackage{fontawesome5}  

\usepackage{hyperref}

\usepackage[table]{xcolor}
\usetikzlibrary{backgrounds,shadows.blur,fit}
\usetikzlibrary{matrix,shapes,arrows,fit,tikzmark}

\newcolumntype{L}[1]{>{\raggedright\let\newline\\\arraybackslash\hspace{0pt}}m{#1}}
\newcolumntype{C}[1]{>{\centering\let\newline\\\arraybackslash\hspace{0pt}}m{#1}}
\newcolumntype{R}[1]{>{\raggedleft\let\newline\\\arraybackslash\hspace{0pt}}m{#1}}

\usepackage[T1]{fontenc}
\usepackage[utf8]{inputenc}
\usepackage{inconsolata}
\usepackage{microtype}
\usepackage{soul}

\definecolor{lightergray}{RGB}{230,230,230}
\definecolor{DarkRed}{RGB}{130,25,0}
\definecolor{DarkGreen}{RGB}{30,130,30}
\definecolor{DarkBlue}{RGB}{0,0,250}
\definecolor{purple}{rgb}{0.5,0,1}
\definecolor{dcyan}{rgb}{0.2,0.6,0.5}
\definecolor{darkgreen}{rgb}{0,200,0}
\definecolor{light-gray}{gray}{0.95} 
\definecolor{darkred}{RGB}{200,0,0}
\definecolor{lightgreen}{RGB}{231,255,219}
\definecolor{lightred}{RGB}{252,231,234}
\definecolor{lightyellow}{RGB}{250,253,191}

\newcommand{\greentext}[1]{\colorbox{lightgreen}{#1}\xspace}

\newcommand{\cmark}{\textcolor{DarkGreen}{\ding{51}}}
\newcommand{\xmark}{\textcolor{red}{\ding{55}}}%

\title{Revisiting Generalization Across Difficulty Levels: \\ It’s Not So Easy}

\author{
Yeganeh Kordi\textsuperscript{$\clubsuit$} \;\;
Nihal V. Nayak\textsuperscript{$\diamondsuit$} \;\; 
Max Zuo\textsuperscript{$\clubsuit$} \;\;
\textbf{Ilana Nguyen}\textsuperscript{$\clubsuit$} \;\;
\textbf{Stephen H. Bach}\textsuperscript{$\clubsuit$}\\
  \textsuperscript{$\clubsuit$}Brown University \; 
  \textsuperscript{$\diamondsuit$}Harvard University \;\\
  {\texttt{\{kordi, stephen\_bach\}@brown.edu}}
  }  
\begin{document}

\maketitle

\input{sections/abstract}
\input{sections/Introduction}
\input{sections/Relatedworks}
\input{sections/cdg_method}
\input{sections/experiments}
\input{sections/findings}
\input{sections/Discussion}
\input{sections/Conclusion}
\input{sections/Limitations}



\bibliography{main}

\clearpage
\appendix
\input{sections/Appendix}

\end{document}

%% file: math_commands.tex
\usepackage{amsmath,amsfonts,bm}

\def\eqref#1{equation~\ref{#1}}

\def\1{\bm{1}}

\DeclareMathAlphabet{\mathsfit}{\encodingdefault}{\sfdefault}{m}{sl}
\SetMathAlphabet{\mathsfit}{bold}{\encodingdefault}{\sfdefault}{bx}{n}



%% file: sections/abstract.tex
\begin{abstract}
We investigate how well large language models (LLMs) generalize across different task difficulties, a key question for effective data curation and evaluation.  
Existing research is mixed regarding whether training on easier or harder data leads to better results, and whether those gains come on easier or harder test data.
We address this question by conducting a systematic evaluation of LLMs' generalization across models, datasets, and fine-grained groups of example difficulty.
We rank examples in six datasets using the outputs of thousands of different LLMs and Item Response Theory (IRT), a well-established difficulty metric in educational testing.
Unlike prior work, our difficulty ratings are therefore determined solely by the abilities of many different LLMs, excluding human opinions of difficulty.
With a more objective, larger-scale, and finer-grained analysis, we show that cross-difficulty generalization is often limited; training on either easy or hard data cannot achieve consistent improvements across the full range of difficulties.
These results show the importance of having a range of difficulties in both training and evaluation data for LLMs, and that taking shortcuts with respect to difficulty is risky.\footnote{Code and data are available at \url{https://github.com/BatsResearch/cross-difficulty}
}
\end{abstract}

%% file: sections/Introduction.tex
\section{Introduction}
\label{sec:intro}

Can language models trained on data at easier task difficulties generalize to harder tasks, or vice versa?
We term this capability \emph{cross-difficulty generalization}.
While several recent studies have investigated this question from various perspectives, the findings remain mixed and often contradictory.
\citet{hase2024unreasonable} found LLMs demonstrate easy-to-hard generalization: LLMs finetuned on easy data and LLMs finetuned on hard data often performed comparably on hard problems within the same domain.
However, \citet{sun2024easytohard_scalable_alignment} observed that while LLMs fine-tuned with easy data may \textit{not necessarily} generalize to harder data, reward models can.
On the other hand, several works find that LLMs demonstrate hard-to-easy generalization: training LLMs on hard data generalizes better to easier tasks than training on easy data or even on all data \citep{yang-hard2easy, pikus2025hardexamplesneedmaximizing}.
\citet{yang-hard2easy} found that training on hard data improves a model's consistency in its ability to solve easier problems more than if it were trained on easier data and tasked to solve harder problems.
\citet{pikus2025hardexamplesneedmaximizing} showed training on the hardest examples using GRPO~\citep{shao2024grpo} consistently outperforms using all data.
Contrary to these works,~\citet{ding2024easy2hard} observed that the best generalization occurs when train and test data have the same level of difficulty.

As shown in Table \ref{tab:difficulty_tension}, despite ongoing research in this area, the relationship between generalization performance and task difficulty remains an open question.
We believe that relying on humans' judgment of difficulty~\citep{hase2024unreasonable, yang-hard2easy, sun2024easytohard_scalable_alignment, ding2024easy2hard} can be one source of uncertainty in the literature.
It also limits the scalability and resolution of difficulty assessment.
In this paper, we systematically evaluate the extent to which LLMs exhibit cross-difficulty generalization, where difficulty is estimated based on the models' observed abilities.

Understanding the prevalence of cross-difficulty generalization is crucial for effective data curation and evaluation of LLMs.
If performance on hard tasks can be improved by training only on easy tasks, existing datasets might already be sufficient to extend language models' capabilities beyond what they currently demonstrate.
Conversely, limited cross-difficulty generalization would place greater importance on curating the right mix of examples based on difficulty during training.

\definecolor{easy}{RGB}{34, 197, 94}      
\definecolor{hard}{RGB}{239, 68, 68}      
\definecolor{local}{RGB}{59, 130, 246}    

\tikzset{   
        every picture/.style={remember picture,baseline},
        every node/.style={anchor=base,align=center,outer sep=1.5pt},
        every path/.style={thick},
        }

\newcommand\marktopleft[2]{%
    \tikz[overlay,remember picture] 
        \node (marker-#1-a) at (.1em,#2) {};%
}
\newcommand\marktopright[1]{%
    \tikz[overlay,remember picture] 
        \node (marker-r) at (-.15em,.3em) {};%
}
\newcommand\markbottom[2]{%
    \tikz[overlay,remember picture] 
        \node (marker-#1-b) at (.1em,#2) {};%
    \tikz[overlay,remember picture,inner sep=3pt]
        \node[draw=#1, draw opacity=0.6, rounded corners,fit={(marker-#1-a.north west) (marker-#1-a.west |- marker-#1-b.south) (marker-r.east |- marker-#1-a.north)}, fill=#1, fill opacity=0.2] {};%
}

\begin{table*}[t]
\centering

\small
\renewcommand{\arraystretch}{1.5}
\setlength{\tabcolsep}{4pt}
\begin{center}
\begin{tabular}{l
                @{\hspace{10pt}}
                L{0.49\textwidth}
                p{0.12\textwidth}
                p{0.17\textwidth}
                @{}}
\toprule
\textbf{Paper} & \textbf{Core Claim} & \textbf{Difficult for}\par \textbf{whom?}& \textbf{Training Method} \\
\midrule

\makecell[tl]{\marktopleft{easy}{.4em}\citet{hase2024unreasonable}}\marktopright &
 Training on easy data performs almost as well on the hard test set as training on hard data. &
LLM + Human &
\makecell[tl]{SFT, ICL, \\ Linear Probing}\\

\makecell[tl]{\citet{sun2024easytohard_scalable_alignment}}\markbottom{easy}{0.em}&
Training only on easy tasks can outperform training on all tasks. &
Human &
RL\\

\makecell[tl]{\marktopleft{hard}{0.5em}\citet{yang-hard2easy}} &
Hard data improves the model's consistency on similar questions more effectively than easy data. &
Human &
SFT, ICL\\

\makecell[tl]{\citet{pikus2025hardexamplesneedmaximizing}}\markbottom{hard}{0.2em} &
Training on the hardest data performs best. &
LLM &
RL\\

\makecell[tl]{\marktopleft{local}{.3em}\citet{ding2024easy2hard}} &
Training provides generalization to similar difficulties, but this generalization reduces as training difficulty increases.&
LLM + Human &
SFT\\

\makecell[tl]{\textbf{Our Analysis}}\markbottom{local}{-0.5em} &
Training on only hard or easy data fails to generalize to other difficulty levels.
Human-centric difficulty metrics are not well-suited for studying LLMs.&
LLM &
SFT \\

\bottomrule
\end{tabular}
\end{center}

\caption{
\textbf{Comparison of related work on cross-difficulty generalization.} 
Prior work focuses on easy-to-hard or hard-to-easy generalization with difficulty metrics obtained from humans, LLMs, or both. 
Our work focuses on understanding cross-difficulty generalization, i.e., both easy-to-hard and hard-to-easy generalization, using model-based difficulty metrics, and shows that limited easy-to-hard and hard-to-easy generalization occurs in LLMs.
Papers highlighted in \textcolor{easy}{green} focus on the effectiveness of easy training data.
Conversely, \textcolor{hard}{red} papers discuss the gain from hard data.
Papers highlighted \textcolor{local}{blue} find that generalization across difficulty levels is limited.
}

\label{tab:difficulty_tension}
\vspace{\baselineskip}
\end{table*}

A key technical challenge in studying generalization is accurately estimating task difficulty.
Difficulty is not a universally agreed-upon metric; it is inherently relative, subjective, and language model-dependent.
Most existing difficulty estimation methods rely on human-based metrics such as grade level and expert ratings or heuristics such as question length and number of reasoning steps required to answer the task~\citep{hase2024unreasonable,sun2024easytohard_scalable_alignment}. 
However, LLMs can struggle in tasks that are considered easy for humans (e.g., counting), and human-based metrics may not capture the difficulty of a task for LLMs correctly.
Some other works proposed using LLM-based metrics such as perplexity, confidence, success rate, and loss~\citep{swayamdipta2020dataset, hase2024unreasonable, pikus2025hardexamplesneedmaximizing}.
However, these approaches usually rely on a small number of LLMs, and the confidence or accuracy of an individual model can be miscalibrated, as models often cannot accurately estimate their own capabilities.
Therefore, analyzing responses and patterns across many LLMs provides a more robust understanding of difficulty. 

\vspace{-1pt}
In this work, we use Item Response Theory (IRT) \citep{baker2001basics, lord1968statistical} to study cross-difficulty generalization.
IRT is a tool widely used to quantify the difficulty of questions and the capability of students in standard educational tests~\citep{kingston1982gre, mckinley1987gre, cook1985sat}.
Here, we treat large language models as the students.
Some metrics for difficulty, such as question or correct response length, examine solely the features intrinsic to benchmark problems themselves. 
Other model-centric metrics, such as model loss or average model accuracy, typically only incorporate model performance on one question at a time.
IRT jointly optimizes question difficulty and LLM ability to better understand the role each plays in model performance.
To estimate the IRT parameters, such as task difficulty and the ability of LLMs, we need to evaluate their performance on the target tasks.
Given that running inference with thousands of models and evaluating them would be prohibitively expensive, we collect existing evaluation results for thousands of large language models from the Open LLM Leaderboard~\citep{open-llm-leaderboard-v1, open-llm-leaderboard-v2}, a popular hub in the community for benchmark results.
After collecting evaluation results, we use IRT to estimate LLM-based difficulty scores for each of six datasets (see \S\ref{cross_diff:irt}).
We find that human-based difficulty metrics diverge substantially from IRT difficulty scores, which are calculated based on LLM abilities, underscoring the value of accurate LLM-based difficulty estimation (\S\ref{cross_diff:correlation}).
Finally, we divide each dataset into ten equal-sized bins, ordered by increasing difficulty, to systematically study generalization. 

\vspace{-1pt}
We train LLMs on each difficulty bin individually and evaluate them on all the other difficulty bins to characterize generalization.
Our experiments show that cross-difficulty generalization is far from a pattern.
First, we find that training language models on one of the difficulty bins can sometimes generalize to easier or harder bins, but the extent of generalization is limited (\S\ref{subsec:finding1}). 
Neither the easy nor the hard training data tends to achieve consistent generalization across the full range of difficulties.
Second, we observe the best generalization between the adjacent bins, and the generalization decreases as we increase the gap between train and test difficulty (\S\ref{subsec:finding2}). 

\vspace{-1pt}
Our main contribution is a comprehensive \textbf{analysis of cross-difficulty generalization} in LLMs: We estimated difficulty scores for each example in six widely used benchmarks and provided a systematic study of generalization across various difficulty levels.
This analysis reveals two key findings:
\begin{itemize}[noitemsep, topsep=0pt]
    \item \textbf{LLMs exhibit weak cross-difficulty generalization:} We demonstrate that for each dataset, across all the models, training solely on easy or hard bins fails to generalize consistently across difficulty levels.
    \item \textbf{Increasing difficulty gap weakens generalization:} We found out that increasing the difficulty gap between the train and test can degrade the performance of the model, sometimes even below the zero-shot baseline.
\end{itemize}

%% file: sections/Relatedworks.tex
\section{Related Work}
\label{sec:related}

\begin{figure*}[!ht]
    \centering
    \includegraphics[width=\textwidth]{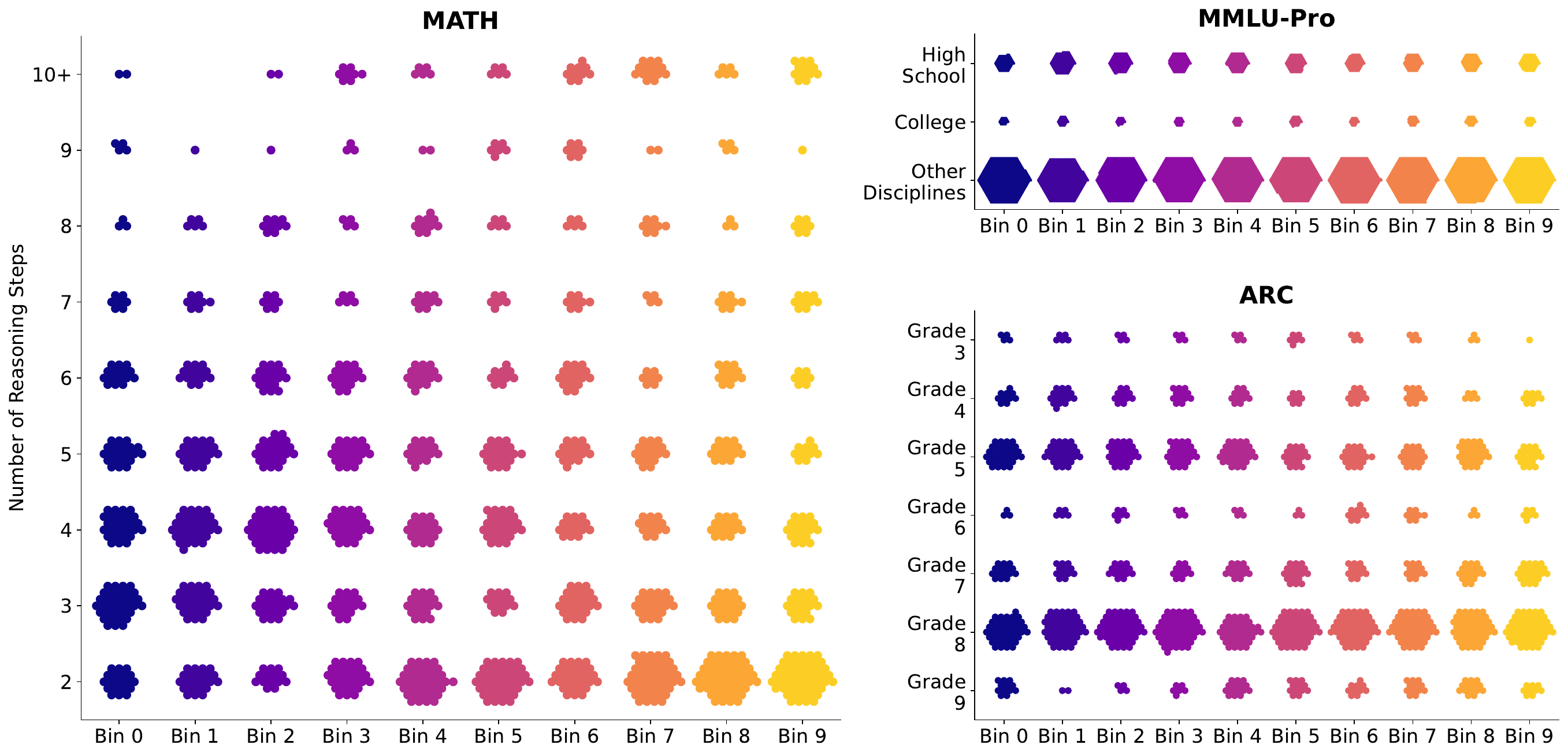}
    \caption{Comparison of human-defined and IRT difficulty estimates for three datasets.
    Each dot represents one question.
    \textbf{Left}: MATH question distribution by number of reasoning steps~\citep{hendrycksmath2021}.
    \textbf{Top right}: MMLU-Pro question distribution by grade level~\cite{wang2024mmlupro}, with questions lacking assigned grades grouped as ``Other Disciplines''.
    \textbf{Bottom right}: ARC question distribution by grade level~\citep{clark2018arc}.
    All distributions are shown across IRT difficulty score bins.}
    \label{fig:correlations-combined-figure}
\end{figure*}

\paragraph{Easy-to-hard Generalization.}

Easy-to-hard generalization is a fast-growing area that studies model generalization on hard tasks by training only on easy examples~\citep{lee2025self_improving,sun2024easytohard_scalable_alignment}.
\citet{hase2024unreasonable} has demonstrated that training on easy data can perform almost as well as training on hard data on hard test sets. 
However, another line of works~\citep{yang-hard2easy, chen-hardReweighted} suggests that training on hard data can show better generalization and consistency compared to using easy data or full data.
For instance,~\citet{pikus2025hardexamplesneedmaximizing} shows that training on the hardest examples yields the largest performance gains, while training on easy examples provides minimal gains.
While these studies highlight an ongoing debate in the community regarding easy-to-hard generalization, their evaluations are often limited to coarse-grained difficulty splits such as easy or hard splits~\citep{hase2024unreasonable,sun2024easytohard_scalable_alignment,pikus2025hardexamplesneedmaximizing}, or focus on rather simple tasks such as digit multiplication~\citep{lee2025self_improving}. 
In our work, we address these limitations by studying cross-difficulty generalization with fine-grained difficulty splits created using item-response theory.

Our work is closely related to \citet{ding2024easy2hard}, but differs in key ways. 
\citet{ding2024easy2hard} primarily benchmarked easy-to-hard generalization using three datasets on the Open LLM Leaderboard and estimated the difficulty of each question using an IRT model with a human-aligned approach.
They then used a greedy search algorithm to select a subset of LLMs whose difficulty rankings best matched human annotators' difficulty judgments.
In contrast, we analyze cross-difficulty generalization using purely model-based difficulty scores, computed directly from model behavior without human calibration.

\paragraph{Difficulty Estimation.}
Understanding what makes examples difficult for language models is critical for robust training and evaluation, and prior work reflects a wide range of approaches, from human annotations to computational metrics.
Many approaches rely on human-based difficulty, using proxies like educational grade level~\citep{clark2018arc, hendrycks2020measuring}, expert ratings~\citep{ding2024easy2hard}, required cognitive skills~\citep{bloom1956taxonomy, hase2024unreasonable}, or indicators such as question length, answer length~\citep{hase2024unreasonable}, and number of reasoning steps to assign difficulty~\citep{fu2023complexity}. 
In Section \ref{cross_diff:correlation}, we show that these metrics do not consistently correlate with each other and model-based difficulties.

Other works suggested using model-based approaches~\citep{swayamdipta2020dataset, ethayarajh22a, varshney2022ildae, voita2020information, perez2021rissanen, ding2024easy2hard} to estimate the difficulty of questions for LLMs. 
\citet{swayamdipta2020dataset} uses training dynamics to estimate the difficulty of data points and provides insights about data quality.
However, it is computationally expensive and requires training the LLM, limiting its scalability.
\citet{ethayarajh22a} propose an information-theoretic approach using V-usable information and pointwise V-information (PVI) to measure question difficulty for a particular model. 
Despite offering more interpretability, calculating V-usable information is also computationally intensive and reflects how informative the data are for a given model class rather than the intrinsic difficulty of the task.
\citet{varshney2022ildae} analyzes difficulty and helps identify trivial or mislabeled examples; however, it relies on model confidence as a proxy for difficulty, which can misrepresent difficulty when a model’s confidence does not estimate its actual ability correctly.
Overall, these methods are either computationally expensive, rely on a limited set of models, or fail to fully reflect the task difficulty as perceived by LLMs.
Instead, our approach uses responses from thousands of LLMs on existing benchmarks to efficiently estimate the difficulty of questions using IRT.

%% file: sections/cdg_method.tex
\section{Cross-Difficulty Generalization with Item Response Theory}
\label{sec:method}

We describe how we constructed datasets used to study cross-difficulty generalization with item-response theory (IRT).
First, we describe the IRT model we use to estimate the difficulty of an example in a dataset (\S\ref{cross_diff:irt}).
Next, we outline a scalable method for collecting evaluation data from a large number of language models, which is then passed to the IRT model to estimate example difficulty (\S\ref{cross_diff:data_collection}). 
Next, we validate the IRT difficulty scores with the zero-shot performance of several large language models (\S\ref{cross_diff:diff_validation}). 
Finally, we compare the difficulty estimates from IRT with those from existing difficulty metrics~(\S\ref{cross_diff:correlation}).

\begin{figure*}[t]
    \centering
    \includegraphics[width=\textwidth]{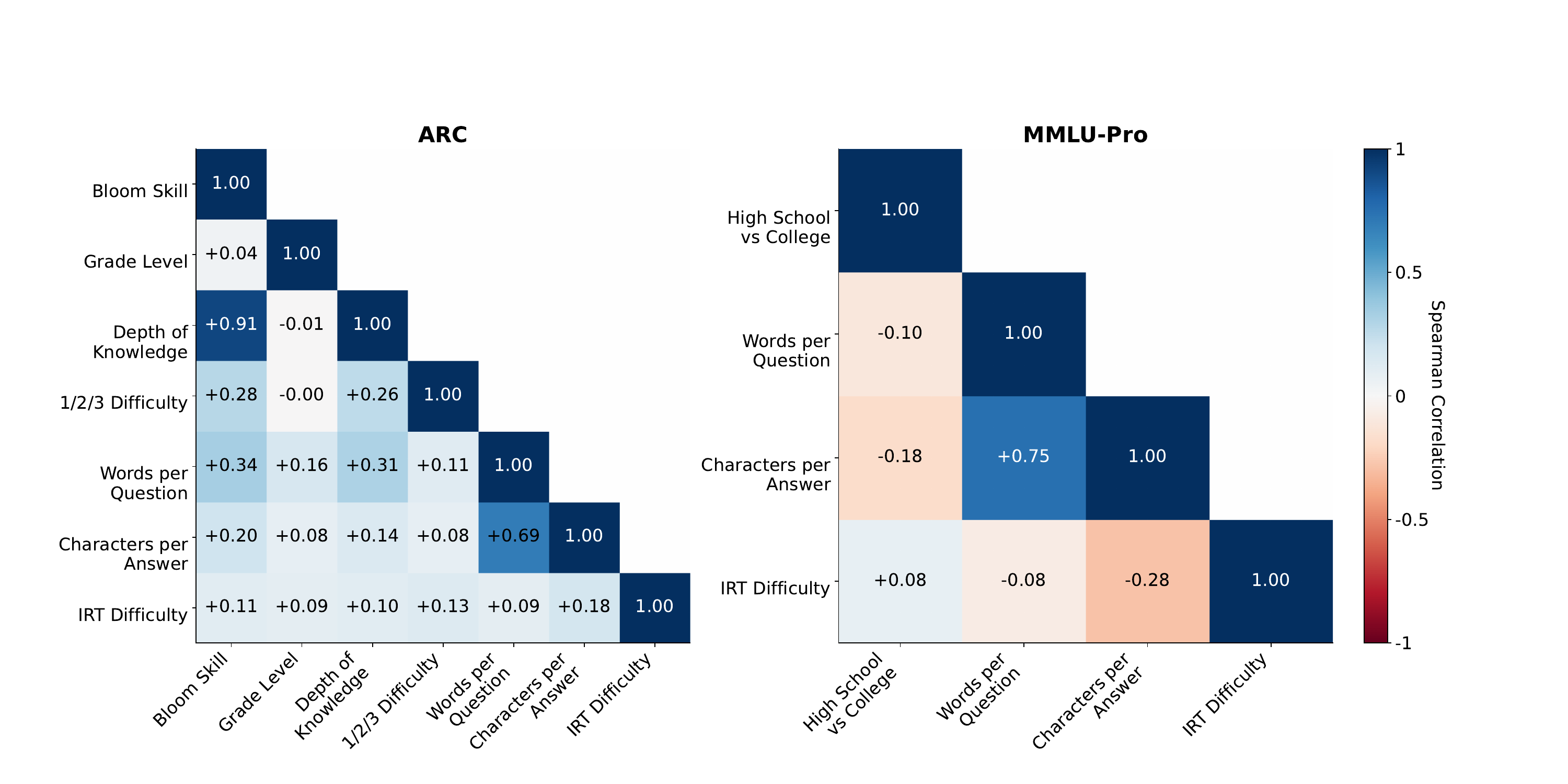}
    \caption{Heatmaps showing Spearman correlations between IRT difficulty scores and human metrics. 
    Colors indicate correlation strength from negative (red) to positive (blue).
    ARC shows weak positive correlations across all metrics, while MMLU-Pro demonstrates mostly no or negative correlation between IRT difficulty and common human metrics for difficulty.}
    \label{fig:arc_gsm8k_correlation_heatmaps}
\end{figure*}

\begin{figure*}[t]
    \centering
    \includegraphics[width=\textwidth]{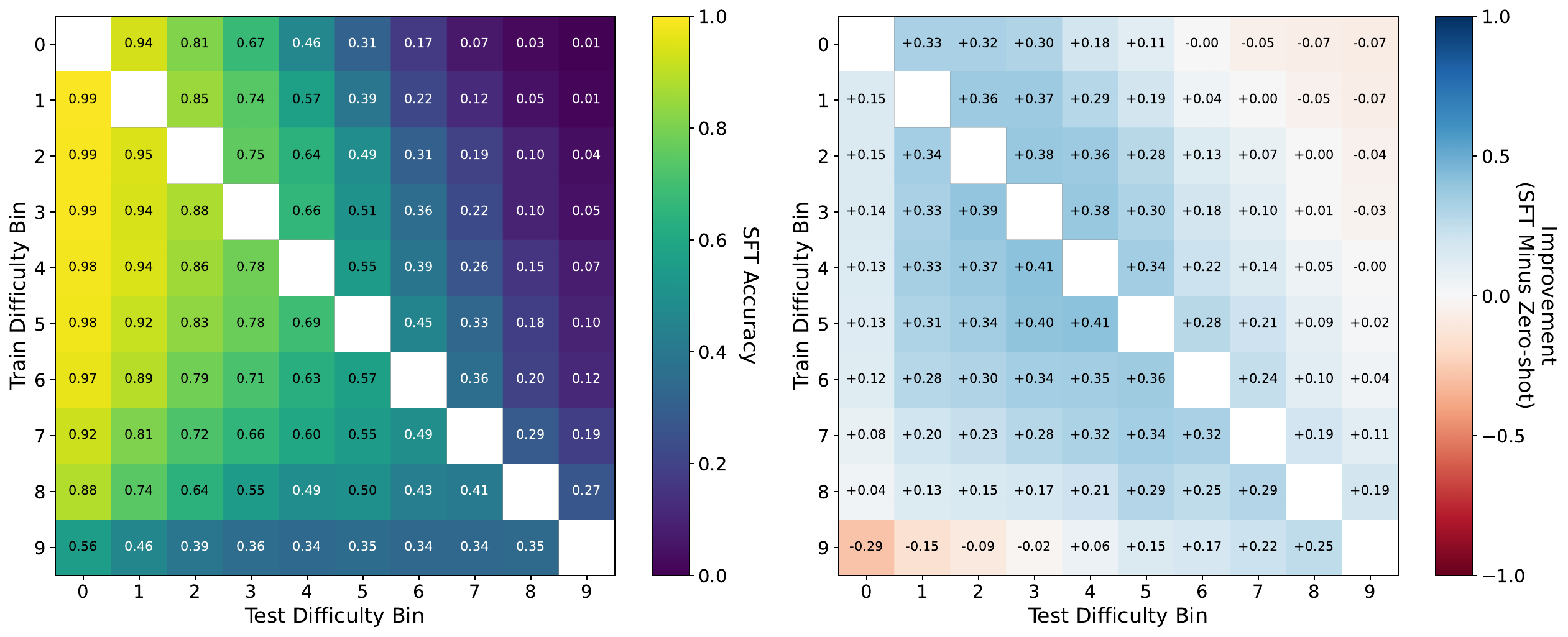}
    \caption{Cross-difficulty generalization heatmaps for Qwen2.5 14B Instruct on MMLU Pro dataset. 
    \textbf{Left}: Performance when training on a difficulty bin (y-axis) and testing on another difficulty bin (x-axis). \textbf{Right}: Improvement from finetuning on each bin compared to the zero-shot performance of the model on that bin.
    Diagonal elements are masked as they represent the same train and test data.}
    \label{fig:qwen14b_heatmaps}
\end{figure*}

\subsection{Difficulty Estimation using IRT}
\label{cross_diff:irt}
To obtain fine-grained difficulty estimates for each example in our datasets, we use Item Response Theory. 
In particular, we use the Rasch (1PL) model~\citep{rasch1960probabilistic} to determine the example's difficulty. 
We are given a dataset $\mathcal{D}=\{x_{1},\dots,x_{N}\}$ with $N$ examples and a set $\mathcal{S}=\{s_{1},\dots,s_{M}\}$ representing $M$ subjects.
In our work, we treat the different language models as the subjects. 
Let $r_{ij}\in\{0,1\}$ be the observed response for the $i$-th task from the $j$-th language model; $r_{ij}=1$ indicates a correct response.
Our goal is to estimate the task difficulty $\{\beta_{i}\}_{i=1}^{N}$ and the ability of the language model $\{\theta_{j}\}_{j=1}^{M}$, that best explain the observed data.
In the 1PL model, the probability of a language model $s_{j}$ correctly answering the task $x_{i}$ is given by: 
\[
P(r_{ij}|\theta_{j},\beta_{i}) = \frac{1}{1+e^{-(\theta_{j}-\beta_{j})}}.
\]
We estimate posterior distributions over the latent parameters of the 1PL model using stochastic variational inference and take their expected values as point estimates.
More details on the 1PL model implementation are provided in the \texttt{py-irt} package by \citet{lalor2023pyirt}.

\begin{table}[t]
    \centering
    \begin{tabular}{lrr}\toprule
         \textbf{Dataset} & \textbf{\# Examples} & \textbf{\# Models}\\\midrule
         ARC & 1,170 & 5,611 \\
         BBH  & 5,250 & 4,354 \\
         GSM8k & 1,319 & 5,870 \\
         MMLU-Pro & 12,032 & 4,359 \\
         MATH & 1,324 & 4,437\\
         MuSR & 756 & 4,433 \\
         \bottomrule
    \end{tabular}
    \caption{\textbf{Statistics for all the datasets.} \# examples is the number of examples from each dataset on Open LLM Leaderboard. \# models is the number of LLMs that has been evaluated on the dataset.\looseness-1}
    \label{tab:dataset_statistics}
\end{table}

\subsection{Data Collection}  
\label{cross_diff:data_collection}
We estimate the difficulties for examples from six datasets: ARC~\citep{clark2018arc}, GSM8k~\citep{cobbe2021gsm}, MMLU-Pro~\citep{wang2024mmlupro}, BBH~\citep{suzgun2023bbh},  MATH~\citep{hendrycksmath2021}, and MuSR~\citep{sprague2024musr} (see more details about the datasets in Appendix \ref{app:datasets}).

We collect responses from the Open LLM Leaderboard~\citep{open-llm-leaderboard-v2, open-llm-leaderboard-v1} to compute difficulties using the IRT model.
Running inference and evaluation with a large number of models can be prohibitively expensive and time-consuming. 
For this reason, similar to~\citet{ding2024easy2hard}, we web scrape the evaluations for all models in the leaderboard to scalably collect responses. 

Table \ref{tab:dataset_statistics} shows the statistics for the datasets.
For each dataset in the leaderboard, we collect responses for all the language models evaluated.
Since the evaluations are typically done on the test set, the number of examples in our dataset corresponds to the test set. 
We also observed that not all the models are evaluated on all tasks. 
Given these responses, we use the IRT model (\S\ref{cross_diff:irt}) to estimate the difficulty of examples in each dataset separately. 
After computing the difficulty scores of each example in the dataset, we sort them and then divide them into ten equally sized bins to study cross-difficulty generalization.

\subsection{Difficulty Validation}
\label{cross_diff:diff_validation}
To further assess the validity of our estimated difficulty scores, we evaluated models from the Qwen3 family~\citep{yang2025qwen3technicalreport}, which were not part of the models used in computing the original IRT-based difficulties.
We conducted zero-shot evaluations across all difficulty bins.
As shown in Figure~\ref{fig:validation} in Appendix~\ref{app:results}, model accuracy consistently decreases with increasing bin difficulty, confirming that the estimated scores generalize reasonably well even to models outside the set used for difficulty estimation.

\subsection{Differences between IRT and Human-Based Metrics}
\label{cross_diff:correlation}
We investigate the correlation between difficulty as measured by IRT and human-based difficulty scores across multiple datasets, revealing key differences between these two approaches.

Following \citet{hase2024unreasonable}, we examined several human-based difficulty metrics.
For all datasets, we examine difficulty indicators, such as answer length (in characters) and question length (in words).
We also include the number of reasoning steps for math datasets (GSM8K \& MATH) and original dataset annotations for school grade/level information for the ARC, MMLU-Pro, and MuSR datasets.
\citet{hase2024unreasonable} additionally labels the ARC dataset with cognitive skill requirements based on Bloom's taxonomy~\citep{bloom1956taxonomy}, Depth of Knowledge~\citep{webb2002dok}, and 1/2/3 expert difficulty ratings.

\vspace{3pt}
The Bloom metric measures the type of cognitive skill a question requires, from basic recall to higher-order reasoning: (1) Remembering, (2) Understanding, (3) Applying, (4) Analyzing, and (5) Evaluating~\citep{bloom1956taxonomy}.
The Depth of Knowledge metric classifies questions by the cognitive process depth needed to arrive at an answer, ranging from Level 1 (Recall and Reproduction) to Level 4 (Extended Thinking)~\citep{webb2002dok}.
Depth of Knowledge complements Bloom by emphasizing task complexity and reasoning depth rather than the type of cognitive skill.
Finally, the 1/2/3 difficulty rating shows expert annotators' overall judgment of question difficulty (easy, medium, hard).

\vspace{2pt}
To visualize these relationships, Figure \ref{fig:correlations-combined-figure} presents each question as a dot plotted against its IRT-derived difficulty score and corresponding human-based metric. 
At each human-based difficulty level, the number of questions in each IRT difficulty bin is roughly uniform, suggesting there is very little correlation between these two approaches.
(See Appendix \ref{app:examples} for some examples from each dataset.)

\vspace{2pt}
We quantify this relationship by calculating the Spearman rank correlation coefficient between each pair of difficulty metrics.
Figure \ref{fig:arc_gsm8k_correlation_heatmaps} shows the Spearman correlation between the IRT and human-based metrics for ARC and MMLU-Pro (See Figure~\ref{fig:correlations_between_hardness_measures} in Appendix \ref{app:corr} for other datasets).

We find that most datasets show very little to no positive correlation between IRT difficulty scores and human-based metrics.
The fact that most human-based metrics weakly correlate with IRT suggests that what makes a task difficult for language models differs from what humans consider a hard question. We include examples of samples that are difficult for LLMs (high IRT scores, incorrect responses across model families) but easy for humans (low human difficulty scores), as well as the reverse in \ref{app:examples}.
Across all datasets and metrics, the highest positive correlations with IRT are observed for answer length in MATH ($0.56$) and the number of reasoning steps in GSM8K ($0.49$).
However, these correlations are inconsistent: the number of reasoning steps negatively correlates with IRT ($-0.08$) in MATH, and answer length negatively correlates with IRT in MMLU-Pro ($-0.28$), MuSR ($-0.13$), IFEval ($-0.14$), and GPQA-Extended ($-0.17$).
Moreover, on average, answer length exhibits little to no positive correlation to most datasets evaluated, contradicting the intuition that longer answers indicate harder problems~\citep{muennighoff2025s1}.

The relatively stronger correlations between IRT difficulty scores and reasoning steps or question length in some datasets show some support for alignment between those human-based metrics and what is actually difficult in practice for LLMs. 
However, the weak correlations for most of the human-based metrics suggest that what makes a task difficult for language models differs from human-based ways of measuring difficulty. 

%% file: sections/experiments.tex
\section{Experimental Setup}
\label{sec:experiments}
We experiment with seven instruction-tuned LLMs from the Qwen 2.5~\citep{qwen2025qwen25technicalreport} and Llama 3~\citep{llama-3} model families.
Here we present the experiments with Qwen2.5 14B Instruct and include experiments and evaluations for Qwen2.5 1.5B/3B/7B Instruct, Llama 3.2 1B/3B, and Llama 3.1 8B in Appendix~\ref{app:results}. 
Across model families and sizes, the results are broadly consistent with the results presented in the main paper.
Implementation details, hyperparameters, and evaluation configurations are described in Appendix~\ref{app:training} and Appendix~\ref{app:evaluation} for full reproducibility.

\paragraph{Training.}
We train the language models using supervised fine-tuning exclusively on a single difficulty bin and repeat this process for all bins in the dataset. 
We format the datasets into instruction-response templates.
All model parameters are trained on the response tokens for five epochs. 
We use HuggingFace transformers and the TRL packages to train the model. 

\paragraph{Evaluation.}
We use \texttt{lm-eval-harness} to evaluate all the datasets. 
We follow the standard evaluation protocols and metrics from \texttt{lm-eval-harness} to evaluate both the zero-shot model and the trained models.
For a given dataset, we compute the accuracy across all difficulty bins (except the training bins).
We report the difference in performance as heatmaps to isolate the effect of cross-difficulty generalization over the zero-shot baseline (see Figure \ref{fig:qwen14b_heatmaps} for a brief illustration).
This evaluation enables us to assess how well a model performs beyond the specific difficulty it was initially exposed to during fine-tuning.

%% file: sections/findings.tex
\begin{figure*}[htbp]
    \centering
    \includegraphics[width=\textwidth]{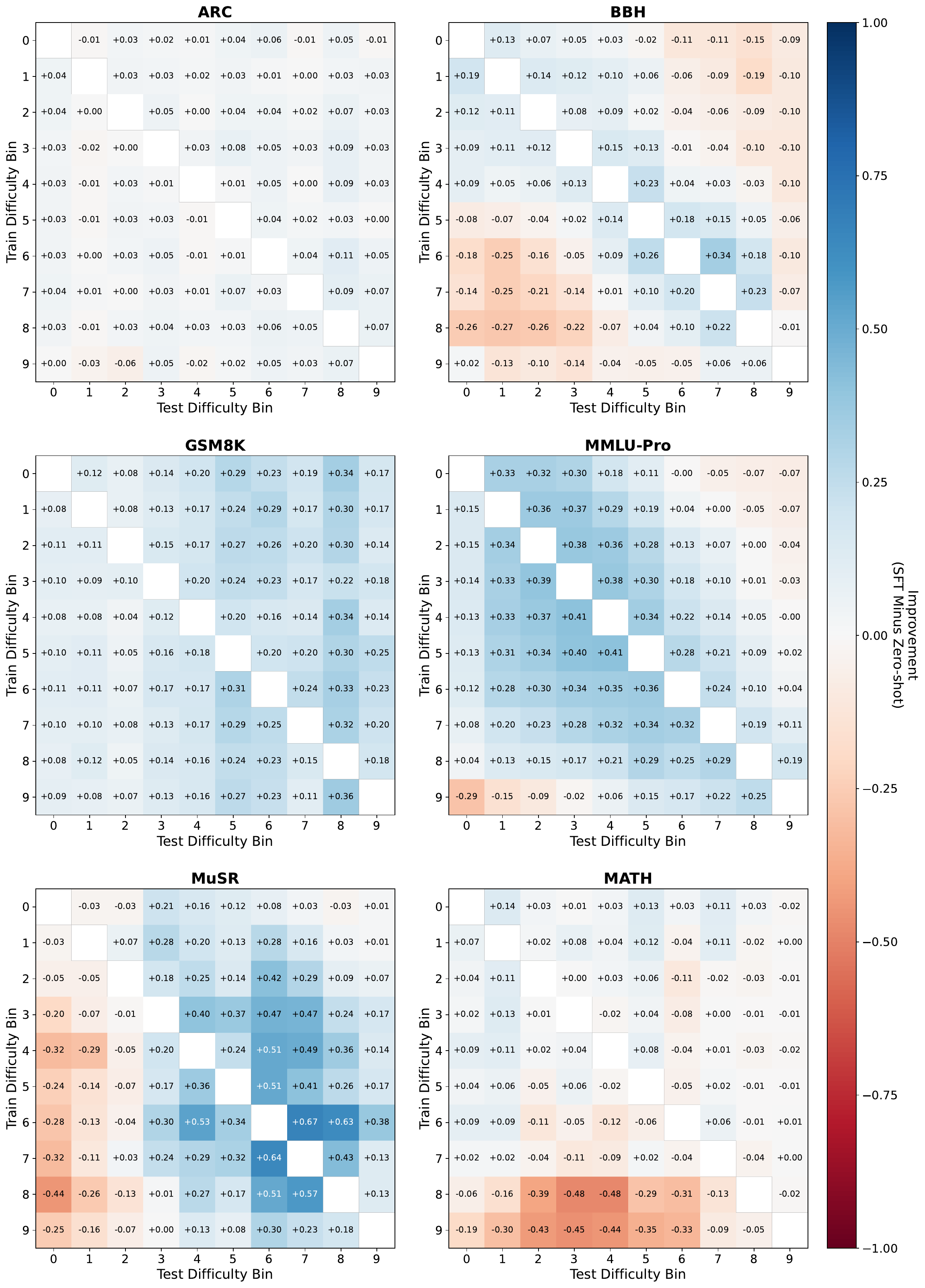}
    \caption{\textbf{Improvement analysis for Qwen2.5 14B Instruct showing the difference between SFT and zero-shot performance}. Blue indicates positive improvements (SFT better than zero-shot), red indicates negative improvements (SFT worse than zero-shot).}
    \label{fig:qwen14b_improvement}
\end{figure*}

\section{Findings}\label{sec:findings}

We report results for Qwen2.5 14B Instruct trained on single difficulty bins and evaluated across all ten bins of six benchmarks.
(See Appendix \ref{app:results} for additional Qwen and Llama models.)
Across settings, we observe limited cross-difficulty generalization, i.e., models trained on easier data fail to generalize consistently to harder data, and those trained on harder data do not generalize consistently to easier data.  
We further find that this generalization capability decreases as the train–test difficulty gap increases, with the strongest performance concentrated near the diagonal, where the difficulties between train and test sets are similar.  
Finally, we observe that these patterns are consistent across model families and sizes, suggesting that they stem from properties of the data distribution rather than from the model.

\paragraph{Easy-to-hard and hard-to-easy generalization are limited.}
\label{subsec:finding1}
Figure \ref{fig:qwen14b_improvement} presents the cross-difficulty generalization results for the Qwen2.5 14B Instruct model.
The rows indicate the training difficulty bins, while the columns represent the test difficulty bins, with both axes ordered from easiest (bin 0) to hardest (bin 9).
The values in each cell are the difference in accuracy between the fine-tuned model and the zero-shot performance.
Positive values suggest good generalization from the corresponding training bin to the test bin, and negative values show poor generalization.
Moving horizontally to the right from each diagonal cell shows easy-to-hard generalization, and moving left shows hard-to-easy generalization.

Our results show that models trained only on easier bins fail to generalize to harder ones.
In the heatmap, this appears as a sharp performance decline when moving right from the diagonal.
For instance, in Figure \ref{fig:qwen14b_heatmaps}, which shows the results for the MMLU-Pro dataset, models trained on the easiest bins (0) perform well on neighboring easy bins but quickly degrade when evaluated on bins five and above.
This pattern directly challenges previous claims that easy-only supervision can recover performance on hard tasks \cite{hase2024unreasonable,sun2024easytohard_scalable_alignment}, while remaining consistent with \citet{hase2024unreasonable}'s observation that supervision gains may decline once the train-test difficulty gap becomes sufficiently large.

We also observe limited hard-to-easy generalization in our results.
In datasets such as BBH, models trained on the hardest bins actually perform worse on easy questions, producing negative values throughout the lower triangle~(Figure~\ref{fig:qwen14b_improvement}). 
This suggests that training exclusively on hard data does not generalize to easier data.

ARC and GSM8K results further demonstrate how cross-difficulty generalization can differ across datasets and model families. 
ARC shows almost no cross-difficulty generalization, with near zero gain across train and test bins. 
GSM8K shows moderate generalization in Qwen2.5 models, suggesting better cross-difficulty generalization, but but this pattern doesn't hold for Llama models (see Appendix \ref{app:results}), where we don't see any generalization for most of the bins.

Overall, the results demonstrate that cross-difficulty generalization is narrow in scope. 
These results challenge the notion that simply training on either easy or hard data can achieve broad generalization across difficulty levels.

\paragraph{Larger train–test difficulty gaps lead to weaker generalization.}
\label{subsec:finding2}
Across datasets, we observe that the strongest generalization values cluster tightly around the diagonal of Figure~\ref{fig:qwen14b_improvement}, where the training and test bins are close in difficulty.
This suggests that models primarily generalize to data of comparable difficulty rather than across significant differences in difficulty.
As the gap between the training and test difficulty increases, accuracy declines, eventually dropping below the zero-shot baseline in both directions.
This pattern of generalization explains why both easy-to-hard and hard-to-easy results drop in Section~\ref{subsec:finding1}, suggesting that the model's ability to generalize is constrained by the difficulty level distribution of its training data rather than by its overall capacity.

\paragraph{Diverse test difficulty is essential.}
When the model is trained on harder examples, we often find that the performance on harder examples improves, but performance on easier examples sees no benefit or even decreases (see Figures~\ref{fig:qwen14b_improvement},~\ref{fig:qwen7b_improvement}, and ~\ref{fig:llama8b_improvement}).
Hence, if a model performs well on challenging benchmark problems, we cannot assume that it also performs well on easy benchmark problems, as we cannot assume hard-to-easy generalization.
This limitation is especially relevant for benchmarks that target only the most challenging problems, such as AIME~\citep{MAA2025AIME} and HLE~\citep{phan2025humanitysexam}, and provide limited information about models' behavior on easier questions.
Therefore, as language models are trained to solve increasingly complex problems, it is essential to continue benchmarking models' capabilities across a broad spectrum of difficulties.

\paragraph{Patterns are consistent across models and datasets.}
In Appendix \ref{app:results}, we show that the limited cross-difficulty generalization patterns persist across model sizes and families.
We evaluate instruction-tuned LLMs ranging from the 1B-parameter to the 14B-parameter in Qwen and Llama model families and observe similar patterns of weak generalization across the models as seen in Figure \ref{fig:qwen14b_improvement}. 
While larger models achieve higher accuracy in absolute numbers, they exhibit weak cross-domain generalization compared to their zero-shot performance.
The consistency in results suggests that our findings will hold across different model sizes. 
\looseness-1

%% file: sections/Discussion.tex
\section{Discussion}

We revisit the growing claims of cross-difficulty generalization in LLMs.
Our study show that cross-difficulty generalization is limited and reduces as we increase the gap between the difficulty of train and test data.  
Generalization is highest where training and testing difficulties are similar, and this pattern remains consistent across model families, scales, and datasets.  

These results suggest that prior works~\citep{yang-hard2easy,hase2024unreasonable, sun2024easytohard_scalable_alignment, pikus2025hardexamplesneedmaximizing} may have overestimated easy-to-hard and hard-to-easy generalization.
Several factors may explain this difference.  
First, we quantify difficulty using IRT and based on models' capabilities and performance rather than human-based metrics or heuristics.  
As shown in Figure~\ref{fig:correlations-combined-figure} and Section~\ref{cross_diff:correlation}, human-based difficulty metrics don't correlate with the IRT results. 
The hardest bin for models may still contain some of the easiest questions for humans, meaning that human-based metrics fail to properly isolate difficulty and produce overlapping distributions.  
Second, we analyze ten distinct difficulty bins rather than two or three splits.  
This allows us to train on one difficulty level and evaluate on problems significantly harder or easier than the training data, giving us a better sense of the true extent of cross-difficulty generalization than if we only compared against neighboring difficulty levels.
Finally, we evaluate across six benchmarks that span reasoning, factual recall, and instruction following, thereby reducing dataset-specific biases rather than testing a narrow set of skills~\citep{sun2024easytohard_scalable_alignment}.\looseness-1

Given the limited cross-difficulty generalization in LLMs, training and evaluation datasets should explicitly account for difficulty alongside other desirable properties such as diversity and coverage.  
An open question remains whether a curriculum structured by model-based difficulty, such as IRT scores, can lead to cross-difficulty generalization.
Future work should explore training objectives and selection strategies that explicitly target performance stability across difficulty levels.

Taken together, these findings reveal a gap in how the field conceptualizes generalization.  
Addressing this gap through precise difficulty measurement, systematic analysis, and difficulty-aware data design is essential for developing models that can extend their reasoning beyond the training distributions.

%% file: sections/Conclusion.tex
\section{Conclusion}
\label{sec:conclusion}
We presented an analysis of cross-difficulty generalization in language models using fine-grained, model-based difficulty estimates from Item Response Theory.  
By training and evaluating models from multiple families and sizes across six benchmarks and ten difficulty bins, we showed that cross-difficulty generalization is limited and highly dependent on the gap between training and evaluation difficulty.
We find that, across model families, model scales, and datasets, as we increase the train-test difficulty gap, generalization from both easy-to-hard data and hard-to-easy data decreases significantly.
These findings challenge the common assumption that training an LLM on either easy or hard data can generalize to data with other difficulty levels.  
Our results motivate a reevaluation of how generalization is measured and improved in LLMs.
A systematic, difficulty-aware perspective will be essential for building models that can perform reliably beyond their training distributions.

%% file: sections/Limitations.tex
\section*{Limitations}
Our analysis relies on publicly available benchmarks (ARC, GSM8K, MMLU-Pro, MATH, MuSR, and BBH) and on fine-grained difficulty scores estimated with Item Response Theory.  
All of these datasets are in English, so our conclusions may not directly extend to tasks in other languages or multilingual settings, or to domains where reliable ground-truth labels are hard to obtain.

We estimate difficulty from model response patterns rather than human annotation.  
While this provides a model-centric understanding of difficulty, shifts in model capabilities, data distributions, or evaluation practices could change the difficulty of each task and affect the strength of the observed patterns.

As with most work on open-source LLMs, we cannot guarantee that evaluation questions were entirely unseen during pretraining.  
Undetected data overlap might inflate accuracy and partially reduce true generalization.  
Although our focus on relative difficulty and cross-bin comparisons can help with this concern, eliminating it would require test sets collected after the pretraining cut-offs of all evaluated models.

Finally, we concentrate on single-bin training to isolate difficulty effects.  
This design clarifies the role of difficulty gaps but does not cover all possible training curricula, such as mixtures of bins or adaptive sampling strategies.
Exploring how such curricula interact with fine-grained difficulty remains an important direction for future work.

\section*{Acknowledgments}
The authors would like to thank the anonymous reviewers for their constructive feedback. We especially thank Ellie Pavlick, Chen Sun, Francisco Piedrahita-Velez, Zheng-Xin Yong, Reza Esfandiarpoor, Yik Siu Chan, Zhenke Liu, Tianze Hua, Apoorv Khandelwal, and other members of the BATS research group and AI Superlab at Brown University for their generous feedback on this work. 
This material is based upon work supported by the National Science Foundation under Grant No. RISE-2425380 and Grant No. IIS-2433429. Any opinions, findings, and conclusions or recommendations expressed in this material are those of the author(s) and do not necessarily reflect the views of the National Science Foundation. Disclosure: Stephen Bach is an advisor to Snorkel AI, a company that provides software and services for data-centric artificial intelligence.

%% file: sections/Appendix.tex
\section{Training Configuration}
\label{app:training}

We fine-tuned a series of instruction-tuned models using full-parameter fine-tuning. 
The models include:
\begin{itemize}[noitemsep, topsep=0pt, leftmargin=1.5em]
    \item Llama3.2 1B Instruct
    \item Llama3.2 3B Instruct
    \item Llama3.1 8B Instruct
    \item Qwen2.5 1.5B Instruct
    \item Qwen2.5 3B Instruct
    \item Qwen2.5 7B Instruct
    \item Qwen2.5 14B Instruct 
\end{itemize}
All models were fine-tuned on difficulty-based subsets of datasets described in Appendix~\ref{app:datasets}.

\paragraph{Optimization.}
Training was conducted with a learning rate of \texttt{5e-6} using the \texttt{paged\_adamw\_8bit} optimizer (\(\beta_1 = 0.9\), \(\beta_2 = 0.99\), weight decay = 0.1). 
We used cosine learning rate decay with a 10\% warmup ratio and clipped gradients to a maximum L2 norm of 0.1. 
Each model was trained for 5 epochs with a batch size of 2 per device across 4 GPUs, resulting in an effective batch size of 8. 
Other parameters were kept at defaults.

\paragraph{Precision and Sequence Length.}
Training used mixed precision with \texttt{bf16} enabled and \texttt{fp16} disabled. 
The maximum input sequence length was 4096 tokens. 
All random seeds (Python, NumPy, and PyTorch) were set to 42 for reproducibility.

\paragraph{Infrastructure.}
Experiments were conducted using \texttt{DeepSpeed ZeRO Stage 3} with single-node distributed training via \texttt{accelerate}. 
We trained on heterogeneous clusters using combinations of NVIDIA A100, A6000, B200, and RTX 3090 GPUs, depending on availability. 
All runs used 8 GPUs per job, a single process per GPU, and no gradient checkpointing. 
Host memory was at least 128GB, and CUDA version 12.4 was used.

\paragraph{DeepSpeed Configuration.}
Training was launched using DeepSpeed, and the configuration is summarized below:
\begin{itemize}[noitemsep, topsep=0pt, leftmargin=1.5em]
    \item \textbf{Zero Stage:} 3 
    \item \textbf{Mixed Precision:} bf16
    \item \textbf{Gradient Accumulation Steps:} 1 
    \item \textbf{Gradient Clipping:} 0.1 
    \item \textbf{Offload Parameters:} False
\end{itemize}

\paragraph{Training Dynamics.}
We monitored training dynamics throughout our experiments and confirmed appropriate learning behavior. 
Training loss curves show consistent convergence across all difficulty bins, and final training accuracy averaged 0.84 across models and datasets.

\paragraph{Reproducibility.}
All training scripts, configurations, and requirements files will be released with the code repository to enable exact reproduction of all reported results.

\section{Evaluation Setup}
\label{app:evaluation}

All models were evaluated using \texttt{lm-eval-harness} with \texttt{vLLM} for efficient inference, and models are loaded in \texttt{bfloat16} precision.
We used greedy decoding (\texttt{temperature=0}, \texttt{top\_p=1.0}, \texttt{max\_new\_tokens=256}) and applied each model’s chat template during formatting. 
Metrics were computed using the official harness implementations, reporting mean accuracy across difficulty bins and evaluation tasks. 
Evaluation was performed on A6000 and A100 GPUs with identical configurations to the training setup.

\begin{figure*}[htbp]
    \centering
    \includegraphics[width=\linewidth]{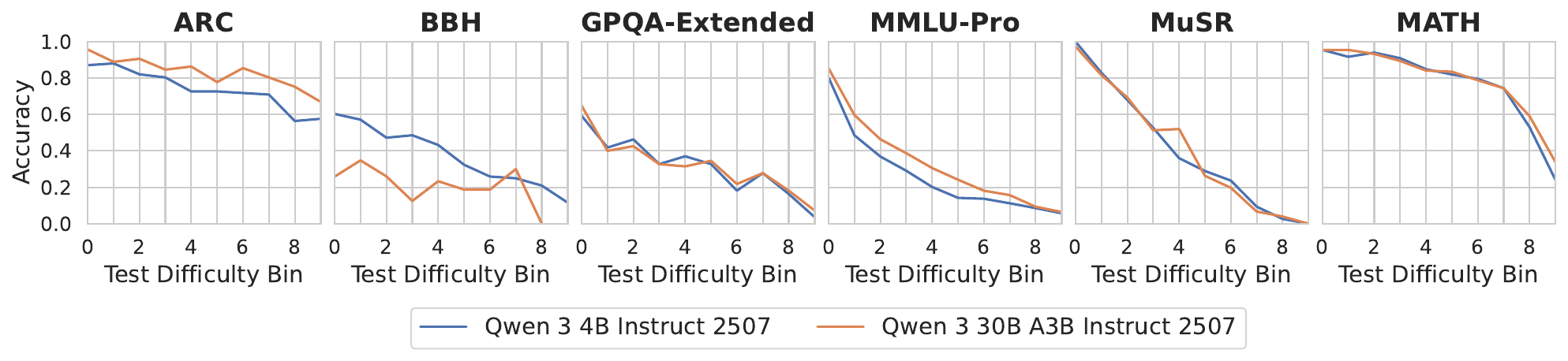}
    \caption{Zero-shot performance of Qwen 3 4B Instruct 2507 and Qwen 3 30B-A3B Instruct 2507 on the same benchmarks we evaluate against, divided by IRT difficulty bins. These models exhibit lower performance on more difficult bins, despite not being calibrated using their model responses.}
    \label{fig:validation}
\end{figure*}

\section{IRT Model Selection}  
In conducting this work, we analyzed multiple IRT formulations, including the 1PL with guessing, 2PL, 3PL, and 4PL models.  
In IRT, these models differ by the parameters they estimate:

\begin{itemize}
    \item 1PL (Rasch): models only question difficulty ($\beta_i$).
    \item 1PL with guessing: adds a fixed guessing lower bound while still modeling only difficulty, allowing a non-zero probability of success even for low-ability models.
    \item 2PL: adds a question discrimination parameter ($\alpha_i$), allowing some questions to separate high- and low-ability models more sharply.
    \item 3PL: introduces a guessing parameter ($c_i$), accounting for a non-zero lower bound (for example, multiple-choice guessing probability).
    \item 4PL: adds an upper asymptote ($d_i$), allowing even the best models to have less than 100\% success probability.
\end{itemize}
While these models introduce additional parameters such as discrimination and guessing, we found that they often led to unstable or counterintuitive estimates because they can explain the same poor model performance with different parameters. 
For example, under the 4PL model, questions that no model could answer were placed near the middle of the difficulty scale (bins 4–5) rather than at the hardest end, despite being infeasible with zero probability of success. 
Similarly, incorporating a guessing parameter caused easy questions to be mislabeled as "guessable" and assigned artificially high difficulty, or vice versa.

To avoid such artifacts, we opted for the simpler 1PL model, which captures difficulty as a single, consistent factor and achieves lower error and more interpretable scores across tasks.

\paragraph{Stability of IRT Difficulty Estimation.}
To further assess robustness, we ran an ablation in which the IRT model was fit using random subsets containing 25\%, 50\%, and 75\% of the available LLMs, and compared the resulting difficulty rankings to those obtained using the full model set.
Across all datasets, the inferred question difficulty ordering exhibits near-perfect agreement with the full-model baseline, with Spearman rank correlations ranging from 0.998 to 1.000.
This shows that reducing the model count does not meaningfully change the relative ordering of question difficulties, indicating that the IRT-based difficulty rankings are stable.

\section{Datasets}  
\label{app:datasets}
To analyze cross-difficulty generalization in language models, we use eight publicly available datasets that span a wide range of reasoning skills and subject areas.  
Together, they cover domains from elementary-level science and math to advanced expert knowledge, instruction following, and specialized reasoning challenges, providing a natural spectrum of task difficulty.  

\paragraph{ARC.} The AI2 Reasoning Challenge (ARC) dataset \cite{clark2018arc} consists of grade-school science questions that test a model’s ability to apply commonsense reasoning and scientific knowledge.  

\paragraph{GSM8K.} GSM8K \cite{cobbe2021gsm} is a collection of 8.5K linguistically diverse grade school math word problems. 
Each problem requires multi-step reasoning and arithmetic calculations, making it a benchmark for evaluating mathematical reasoning.  

\paragraph{MMLU-Pro.} The MMLU-Pro dataset \cite{wang2024mmlupro} is an enhanced version of the Massive Multitask Language Understanding benchmark \cite{hendrycks2020measuring}. 
It assesses LLMs across a broad range of subjects, including STEM, humanities, and social sciences, using multiple-choice questions requiring expert-level knowledge.  

\paragraph{IFEval.} The Instruction Following Evaluation (IFEval) dataset \cite{zheng2023ifeval} measures how well models follow natural language instructions.  
Unlike traditional benchmarks, IFEval is structured as a test set and uses an evaluation algorithm instead of gold labels.  
Because our framework requires training and evaluation on the same set to avoid distribution shift, we constructed gold labels for this dataset by annotating one correct answer for each question.
These labels allow us to integrate IFEval into our pipeline while preserving its role as a challenging test of instruction-following ability.

\paragraph{GPQA-Extended.} The GPQA-Extended benchmark \cite{rein2024gpqa} consists of graduate-level multiple-choice questions across physics, biology, and chemistry. 
The questions are designed to be challenging even for advanced models, making GPQA-Extended a strong test of reasoning capabilities.  

\paragraph{BBH.} The BIG-Bench Hard (BBH) dataset \cite{suzgun2023bbh} is a subset of the BIG-Bench benchmark focusing on tasks that are particularly difficult for LLMs. 
It includes diverse reasoning and comprehension problems that test generalization under challenging conditions.  

These six datasets provide a comprehensive evaluation suite for studying cross-difficulty generalization.  
In the next subsection, we assign fine-grained difficulty scores to each instance in these datasets using Item Response Theory (IRT).  

\paragraph{MATH.}
The MATH dataset \citep{hendrycksmath2021} consists of competition-level problems from high school and undergraduate mathematics domains.
It covers four major categories: Algebra, Number Theory, Counting and Probability, and Geometry.
Problems and solutions are consistently formatted using \LaTeX{}, allowing for the flexible encoding of mathematical expressions.
Following the setting used in the Open LLM Leaderboard, we use only Level~5 problems, which represent the hardest difficulty level in the dataset.

Directly finetuning on the final answers or the provided solutions in the MATH dataset resulted in significant decreases from zero-shot performance for all models.
To mitigate this, we created training samples specific to both Qwen and Llama models.
Specifically, we collect responses from Qwen 2.5 14B and Llama 3.1 8B at several temperatures: 0, 0.1, 0.2, and 0.7, collecting 16 samples for each prompt.
For every correct response, we replace the dataset's given solution with our sampled solution.
We further sample these models with few-shot prompting at higher temperatures (0.7, 1.0, and 1.2), collecting 32 samples for each prompt.
We prompt using three examples, each consisting of: (a) the question, (b) the correct final answer, and (c) a correct response sampled from the model itself.
Any correct response from these prompts also replaced the MATH dataset's provided solution.
This resulted in two datasets: one MATH dataset with solutions that more closely resembled Qwen 2.5 14B style text, and another with solutions that more closely resembled Llama 3.1 8B style text.
We finetune all our Qwen 2.5 and Llama 3.1 models with these datasets.

\paragraph{MuSR.}
MuSR~\cite{sprague2024musr} evaluates multi-step soft reasoning in long-form natural language narratives.
Instances are complex stories (e.g., 1000-word murder mysteries), and models must use clues to answer the questions about the narrative.

\paragraph{Licensing.}
All of these datasets are publicly available, and we use them in accordance with with their official licenses: GSM8K (\href{https://huggingface.co/datasets/openai/gsm8k}{MIT}), ARC (\href{https://huggingface.co/datasets/allenai/ai2_arc}{CC BY-SA 4.0}), MMLU-Pro (\href{https://huggingface.co/datasets/TIGER-Lab/MMLU-Pro}{MIT}), IFEval (\href{https://huggingface.co/datasets/google/IFEval}{Apache 2.0}), GPQA-Extended (\href{https://huggingface.co/datasets/Idavidrein/gpqa}{CC BY 4.0}), BBH (\href{https://github.com/suzgunmirac/BIG-Bench-Hard}{MIT}), MATH (\href{https://github.com/hendrycks/math}{MIT}), and MuSR (\href{https://github.com/Zayne-sprague/MuSR}{MIT}).
To the best of our knowledge, based on the documentation of the datasets and our checks, none of these datasets contain personally identifiable information (PII) or intentionally offensive content.

\paragraph{Label Quality Analysis.}
We manually reviewed samples in the hardest difficulty bin to assess whether label noise contributes to high IRT difficulty scores.
Across datasets, we did not find evidence of systematic mislabeling in the hardest bin.
For example, in GSM8K, we identified five mislabeled questions in the hardest bin, representing less than 4\% of that subset, which is insufficient to explain the observed difficulty distribution.

\section{Supplementary Results}
\label{app:results}
Due to space limitations, the main paper presents results only for the Qwen2.5 14B Instruct model. 
Here, we provide the corresponding heatmaps for the remaining six models.

We also include the zero-shot results for the Qwen3 family discussed in Section \ref{cross_diff:diff_validation}. The full visualization is shown in Figure~\ref{fig:validation}, which complements the main-text analysis.

We also report experiments on the IFEval and GPQA-Extended datasets. 
As shown in Table~\ref{tab:dataset_statistics}, both datasets are relatively small, each containing fewer than 550 examples. 
Consequently, after splitting by difficulty, each fine-tuning bin included fewer than 55 examples. 
Under these conditions, the models didn't show meaningful evidence of learning the tasks, likely due to the limited number of examples, and we observed no consistent patterns of cross-difficulty generalization.
For completeness, we include their results here but exclude them from the main paper.

\begin{figure*}[htbp]
    \centering
    \includegraphics[width=\textwidth]{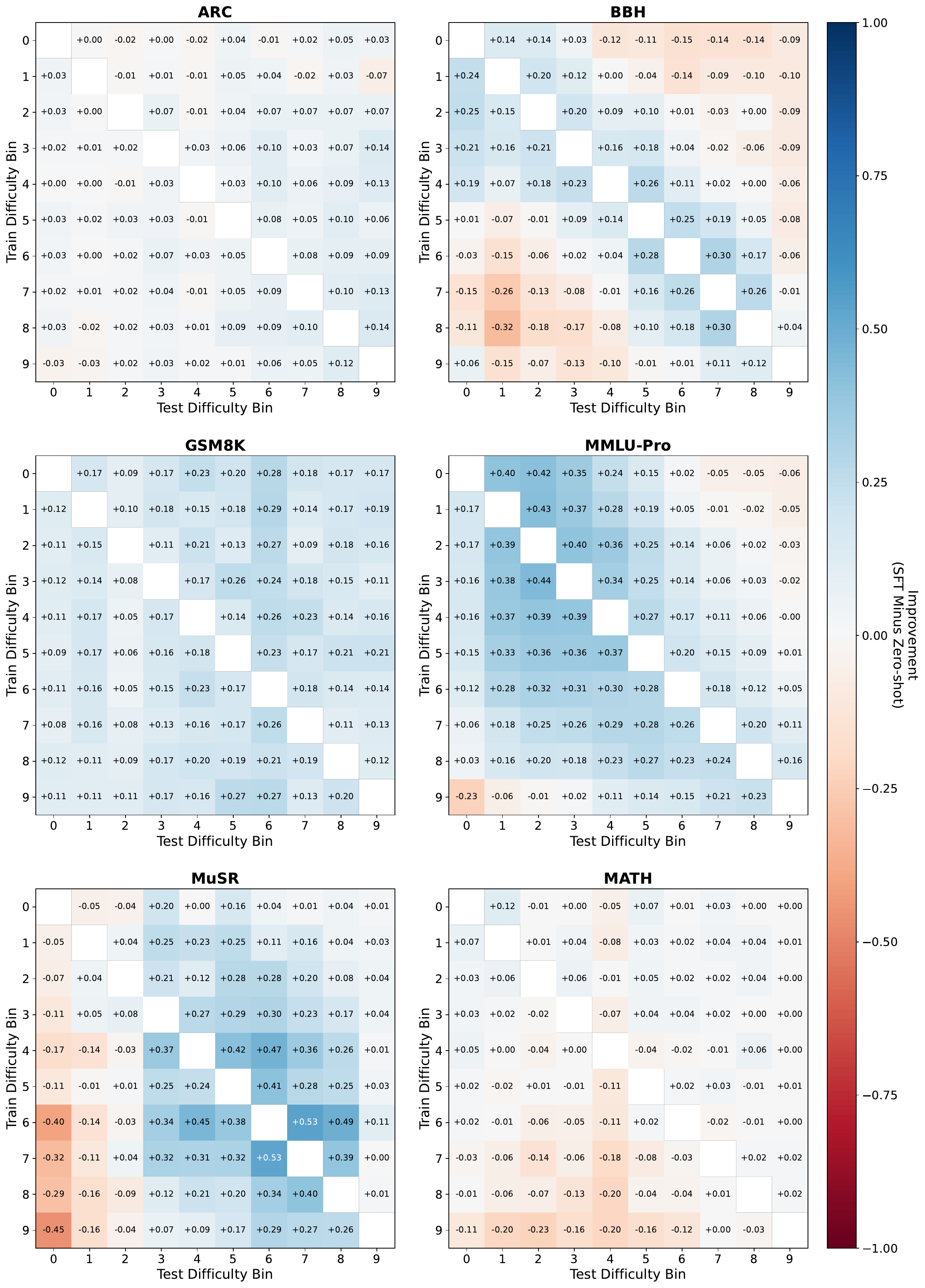}
    \caption{\textbf{Improvement analysis for Qwen2.5 7B Instruct showing the difference between SFT and zero-shot performance.} Blue indicates positive improvements (SFT better than zero-shot), red indicates negative improvements (SFT worse than zero-shot).}
    \label{fig:qwen7b_improvement}
\end{figure*}

\begin{figure*}[htbp]
    \centering
    \includegraphics[width=\textwidth]{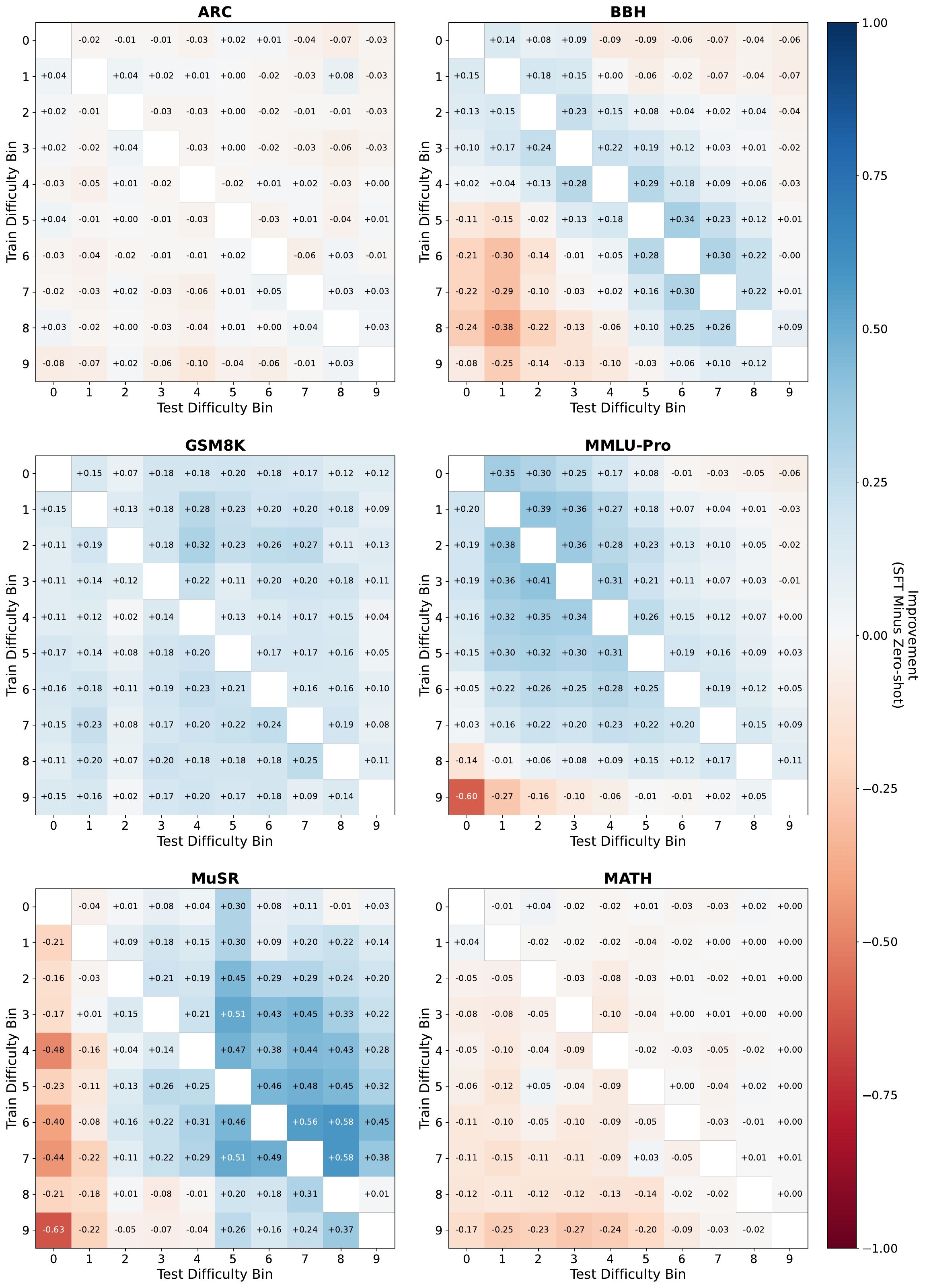}
    \caption{\textbf{Improvement analysis for Qwen2.5 3B Instruct showing the difference between SFT and zero-shot performance.} Blue indicates positive improvements (SFT better than zero-shot), red indicates negative improvements (SFT worse than zero-shot).}
    \label{fig:qwen3b_improvement}
\end{figure*}

\begin{figure*}[htbp]
    \centering
    \includegraphics[width=\textwidth]{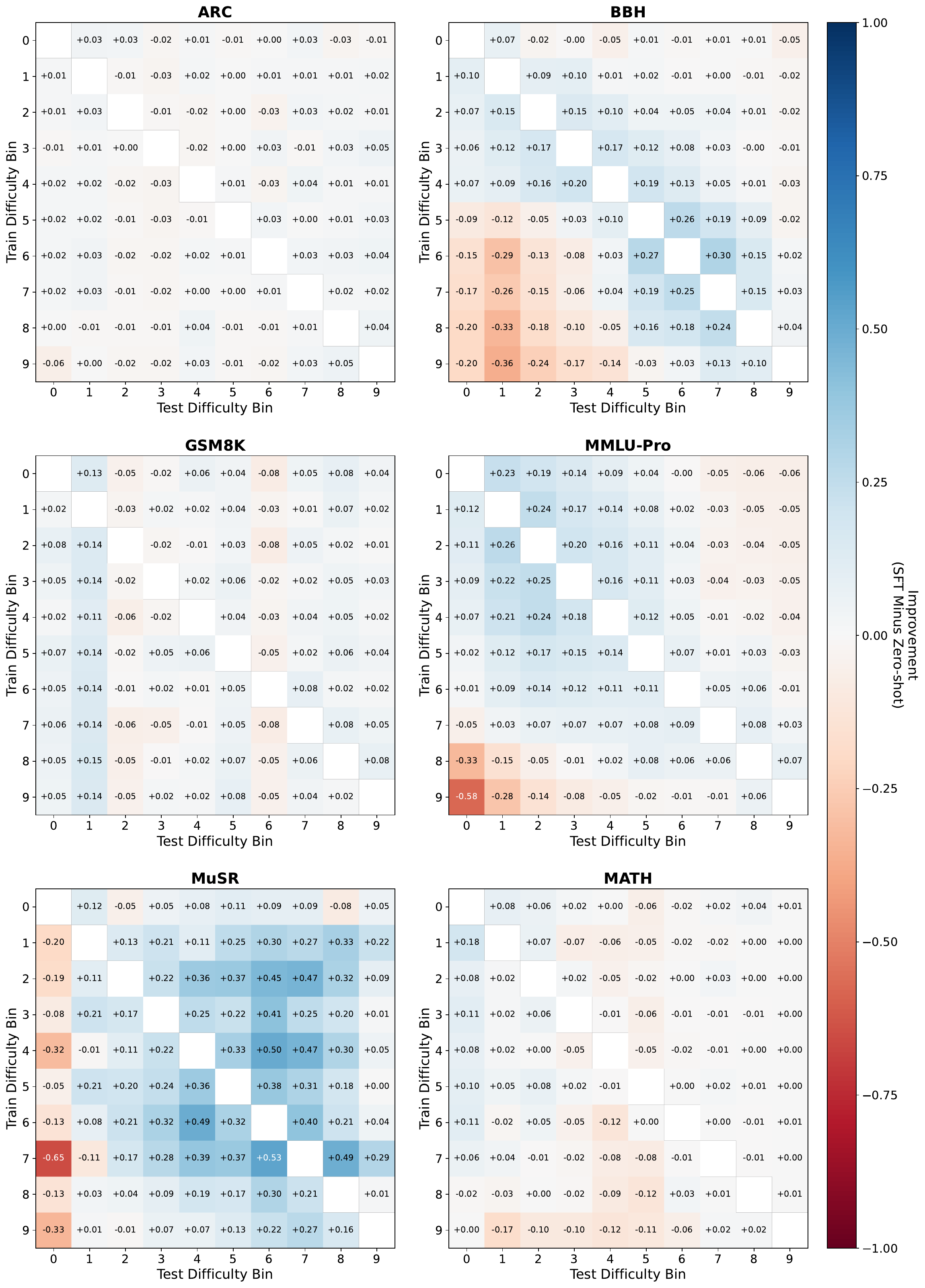}
    \caption{\textbf{Improvement analysis for Qwen2.5 1.5B Instruct showing the difference between SFT and zero-shot performance.} Blue indicates positive improvements (SFT better than zero-shot), red indicates negative improvements (SFT worse than zero-shot).}
    \label{fig:qwen1.5b_improvement}
\end{figure*}

\begin{figure*}[htbp]
    \centering
    \includegraphics[width=\textwidth]{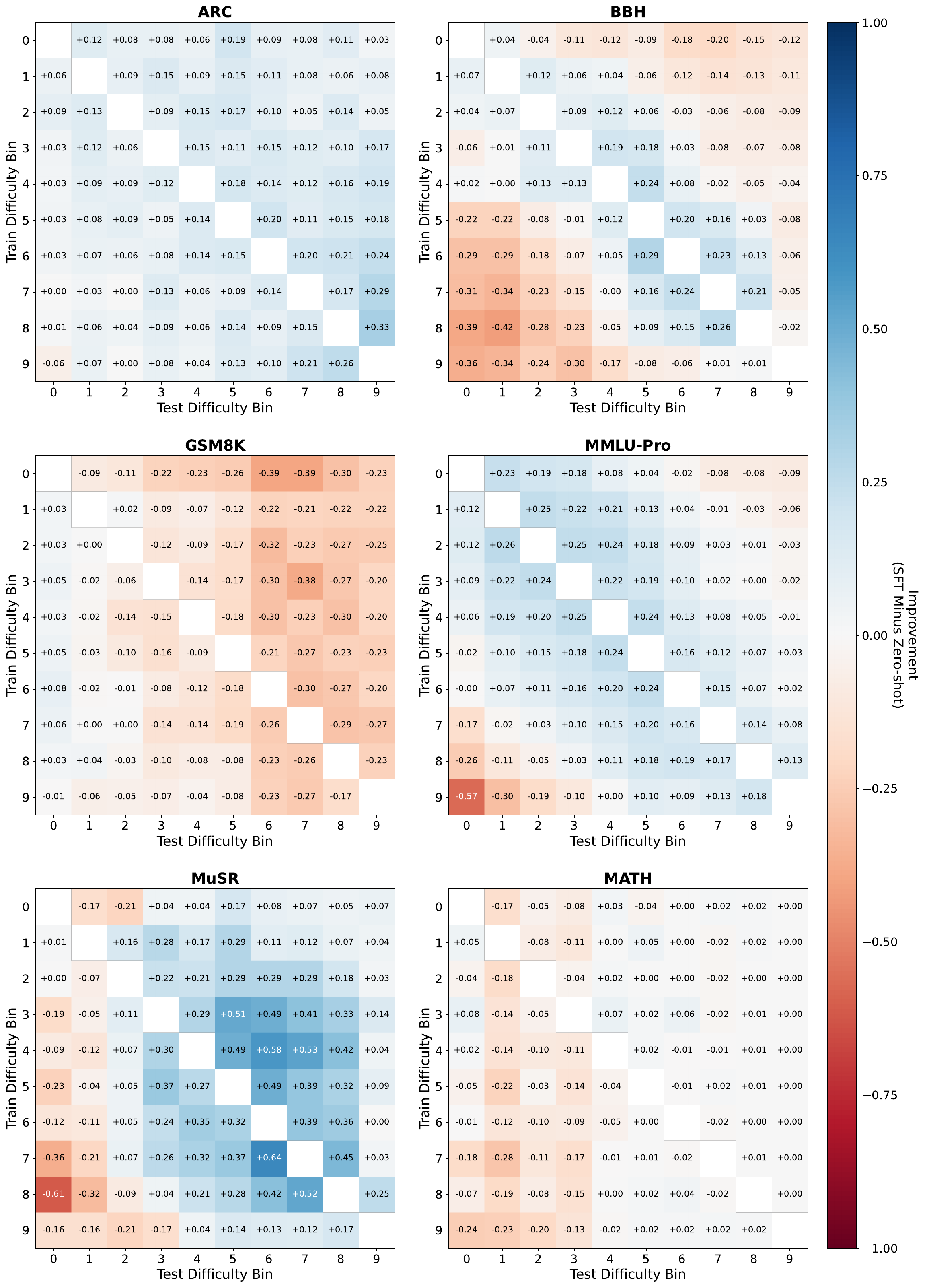}
    \caption{\textbf{Improvement analysis for Llama3.1 8B Instruct showing the difference between SFT and zero-shot performance.} Blue indicates positive improvements (SFT better than zero-shot), red indicates negative improvements (SFT worse than zero-shot).}
    \label{fig:llama8b_improvement}
\end{figure*}

\begin{figure*}[htbp]
    \centering
    \includegraphics[width=\textwidth]{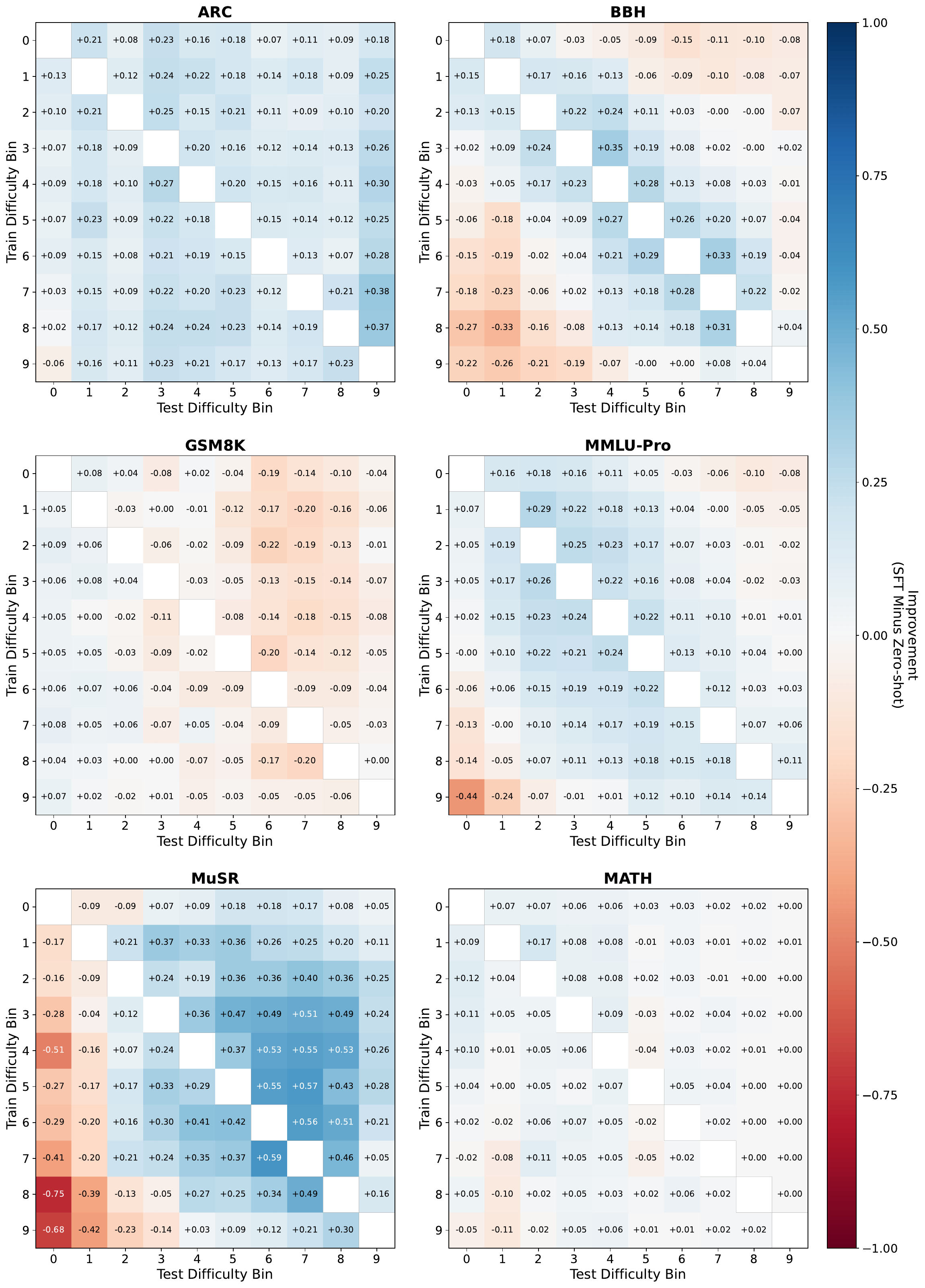}
    \caption{\textbf{Improvement analysis for Llama3.2 3B Instruct showing the difference between SFT and zero-shot performance.} Blue indicates positive improvements (SFT better than zero-shot), red indicates negative improvements (SFT worse than zero-shot).}
    \label{fig:llama3b_improvement}
\end{figure*}

\begin{figure*}[htbp]
    \centering
    \includegraphics[width=\textwidth]{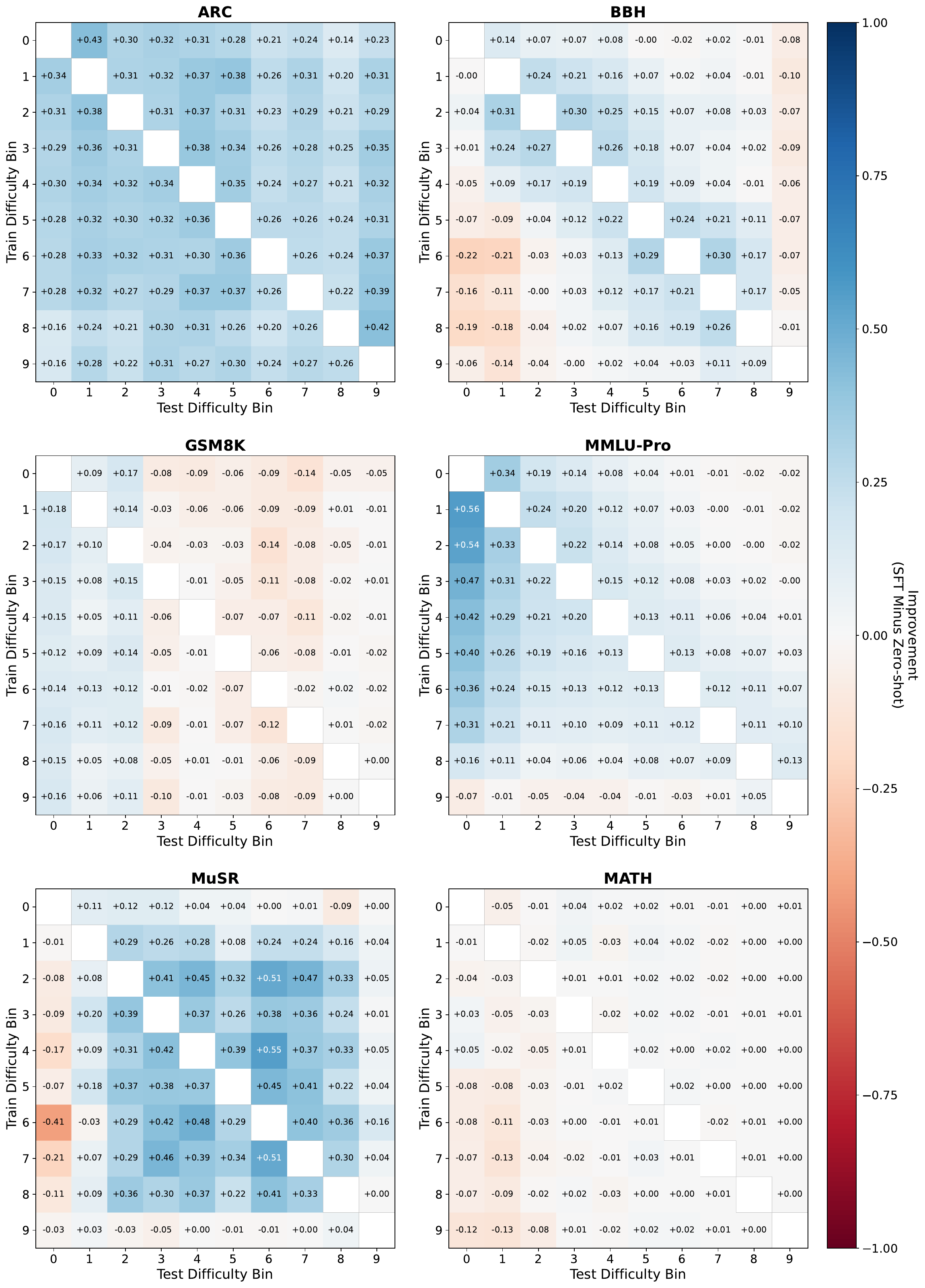}
    \caption{\textbf{Improvement analysis for Llama3.2 1B Instruct showing the difference between SFT and zero-shot performance.} Blue indicates positive improvements (SFT better than zero-shot), red indicates negative improvements (SFT worse than zero-shot).}
    \label{fig:llama1b_improvement}
\end{figure*}

\begin{figure*}[htbp]
    \centering
    \includegraphics[width=\textwidth]{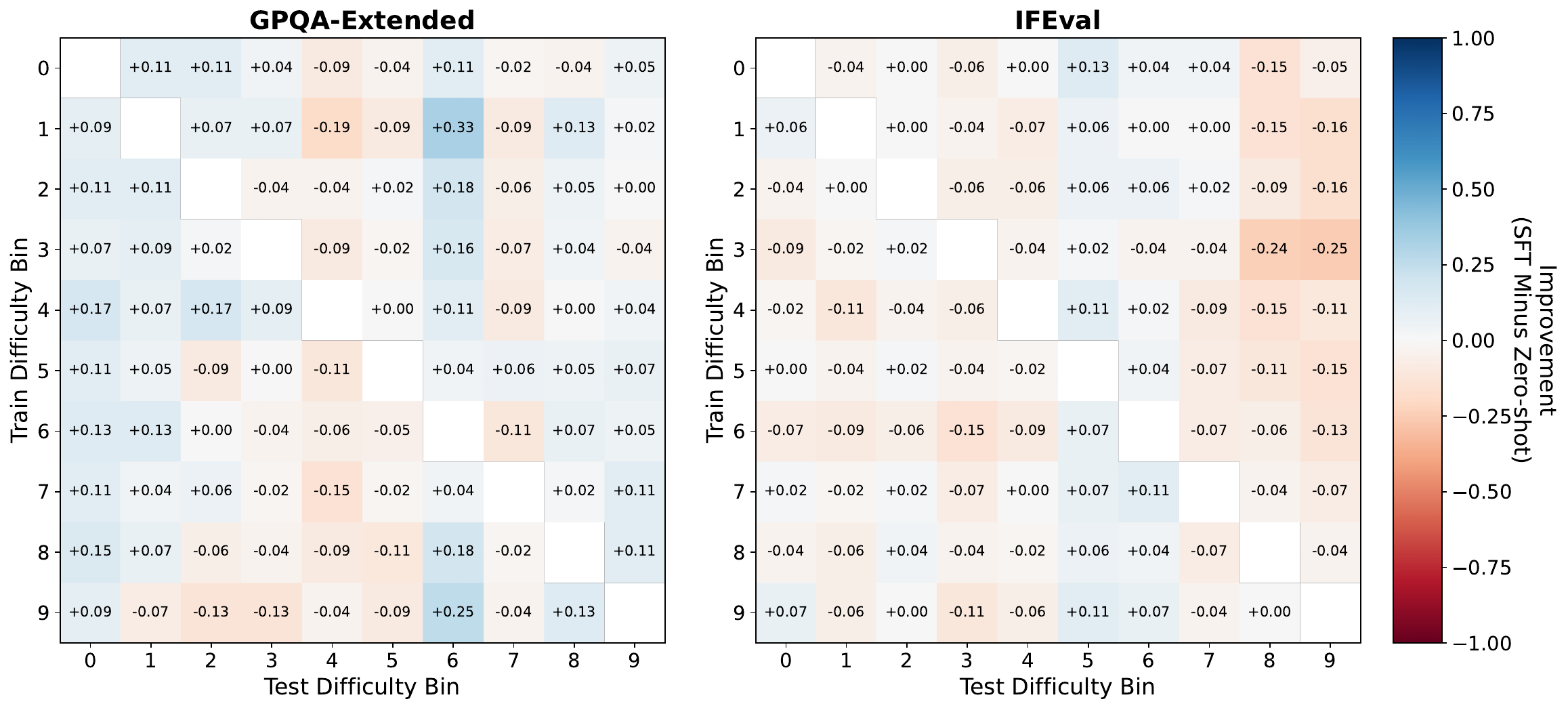}
    \caption{\textbf{Improvement analysis on IFEval and GPQA-Extended for Qwen2.5 14B Instruct showing the difference between SFT and zero-shot performance.} Blue indicates positive improvements (SFT better than zero-shot), red indicates negative improvements (SFT worse than zero-shot).}
    \label{fig:qwen14b_extra}
\end{figure*}

\begin{figure*}[htbp]
    \centering
    \includegraphics[width=\textwidth]{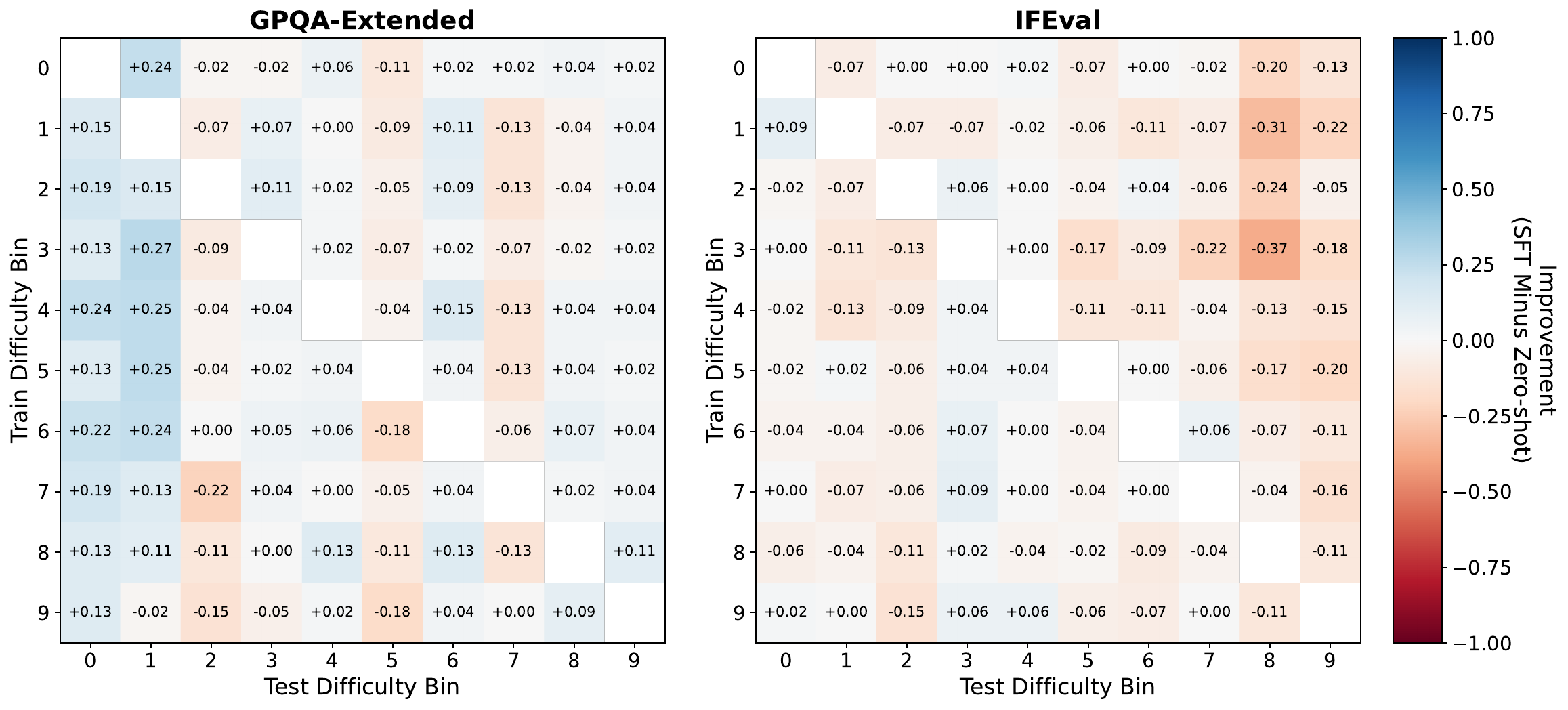}
    \caption{\textbf{Improvement analysis on IFEval and GPQA-Extended for Qwen2.5 7B Instruct showing the difference between SFT and zero-shot performance.} Blue indicates positive improvements (SFT better than zero-shot), red indicates negative improvements (SFT worse than zero-shot).}
    \label{fig:qwen7b_extra}
\end{figure*}

\begin{figure*}[htbp]
    \centering
    \includegraphics[width=\textwidth]{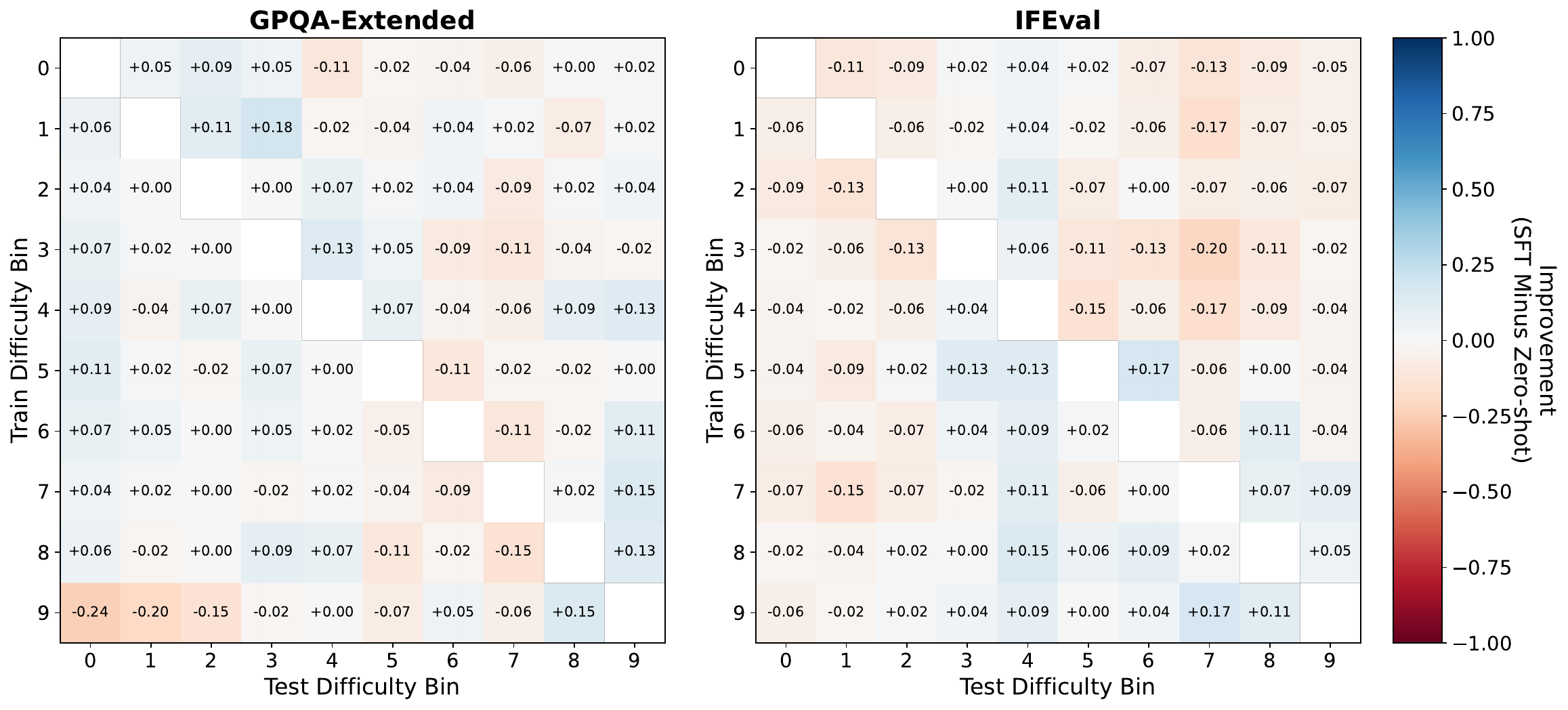}
    \caption{\textbf{Improvement analysis on IFEval and GPQA-Extended for Qwen2.5 3B Instruct showing the difference between SFT and zero-shot performance.} Blue indicates positive improvements (SFT better than zero-shot), red indicates negative improvements (SFT worse than zero-shot).}
    \label{fig:qwen3b_extra}
\end{figure*}

\begin{figure*}[htbp]
    \centering
    \includegraphics[width=\textwidth]{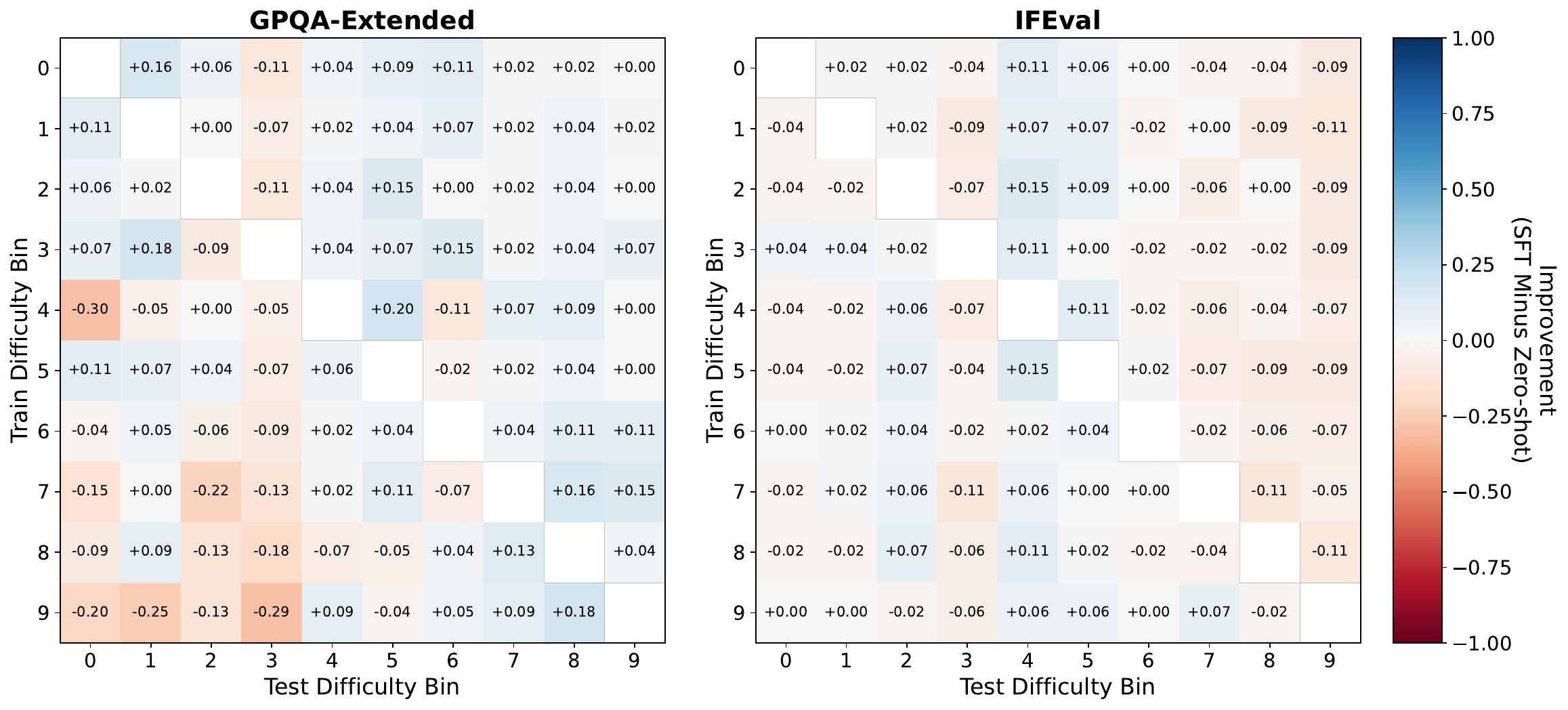}
    \caption{\textbf{Improvement analysis on IFEval and GPQA-Extended for Qwen2.5 1.5B Instruct showing the difference between SFT and zero-shot performance.} Blue indicates positive improvements (SFT better than zero-shot), red indicates negative improvements (SFT worse than zero-shot).}
    \label{fig:qwen1.5b_extra}
\end{figure*}

\begin{figure*}[htbp]
    \centering
    \includegraphics[width=\textwidth]{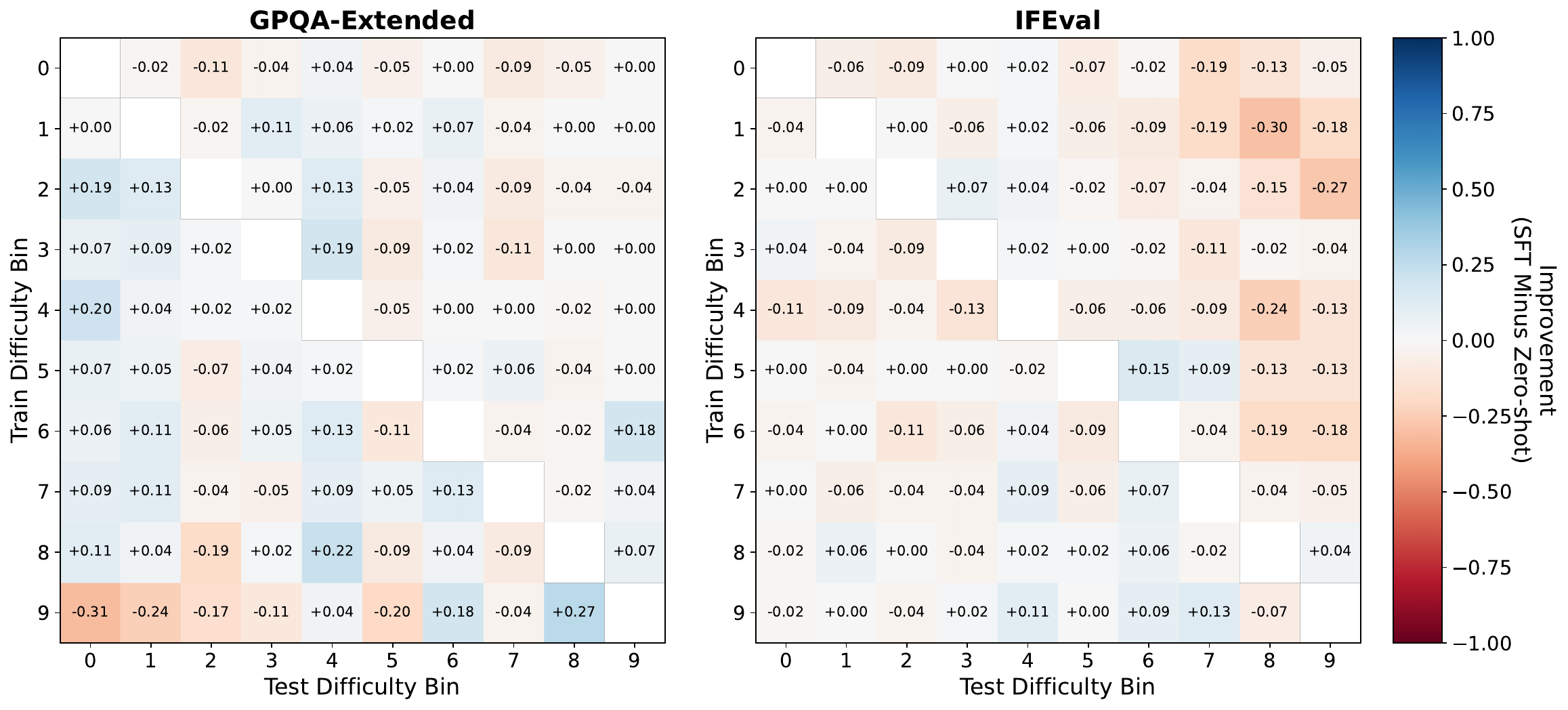}
    \caption{\textbf{Improvement analysis on IFEval and GPQA-Extended for Llama3.1 8B Instruct showing the difference between SFT and zero-shot performance.} Blue indicates positive improvements (SFT better than zero-shot), red indicates negative improvements (SFT worse than zero-shot).}
    \label{fig:llama8b_extra}
\end{figure*}

\begin{figure*}[htbp]
    \centering
    \includegraphics[width=\textwidth]{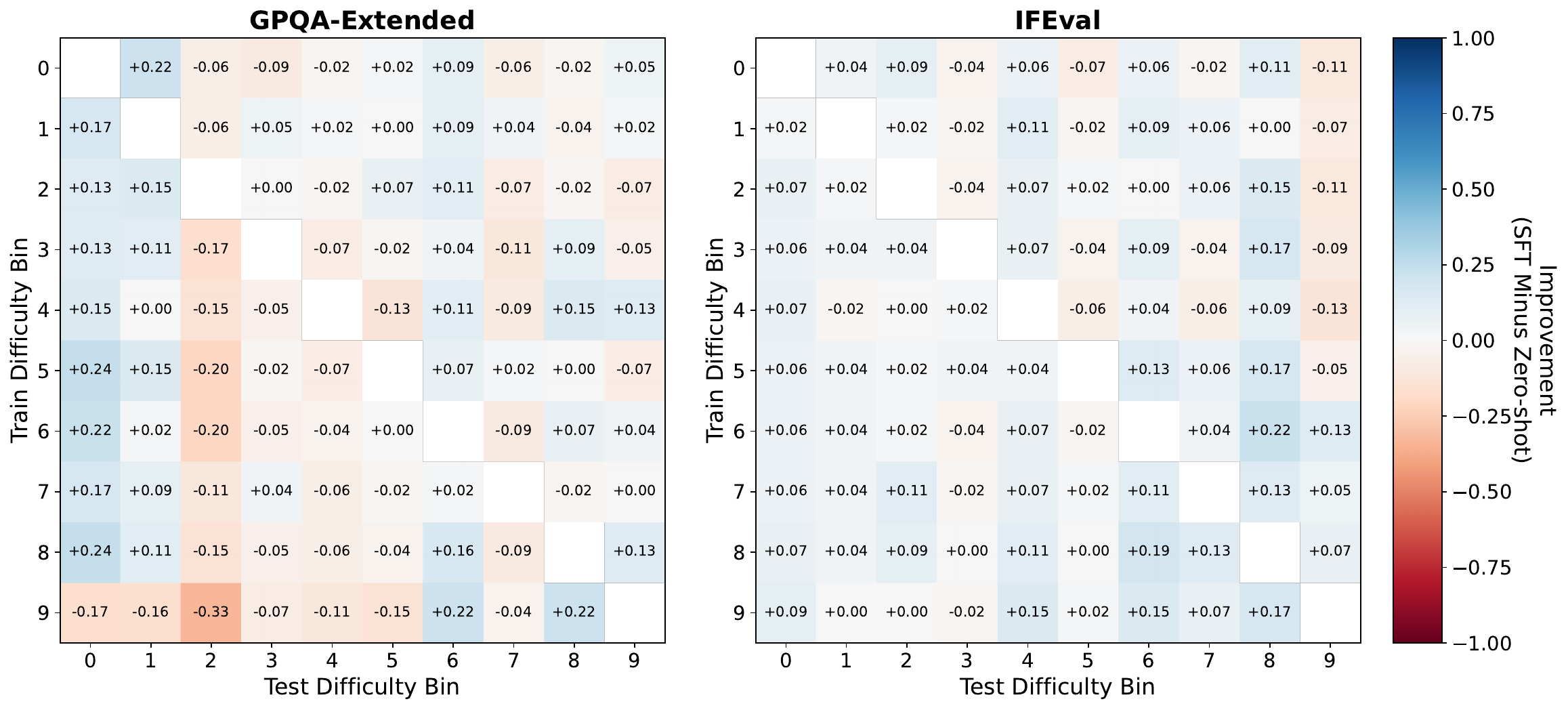}
    \caption{\textbf{Improvement analysis on IFEval and GPQA-Extended for Llama3.2 3B Instruct showing the difference between SFT and zero-shot performance.} Blue indicates positive improvements (SFT better than zero-shot), red indicates negative improvements (SFT worse than zero-shot).}
    \label{fig:llama3b_extra}
\end{figure*}

\begin{figure*}[htbp]
    \centering
    \includegraphics[width=\textwidth]{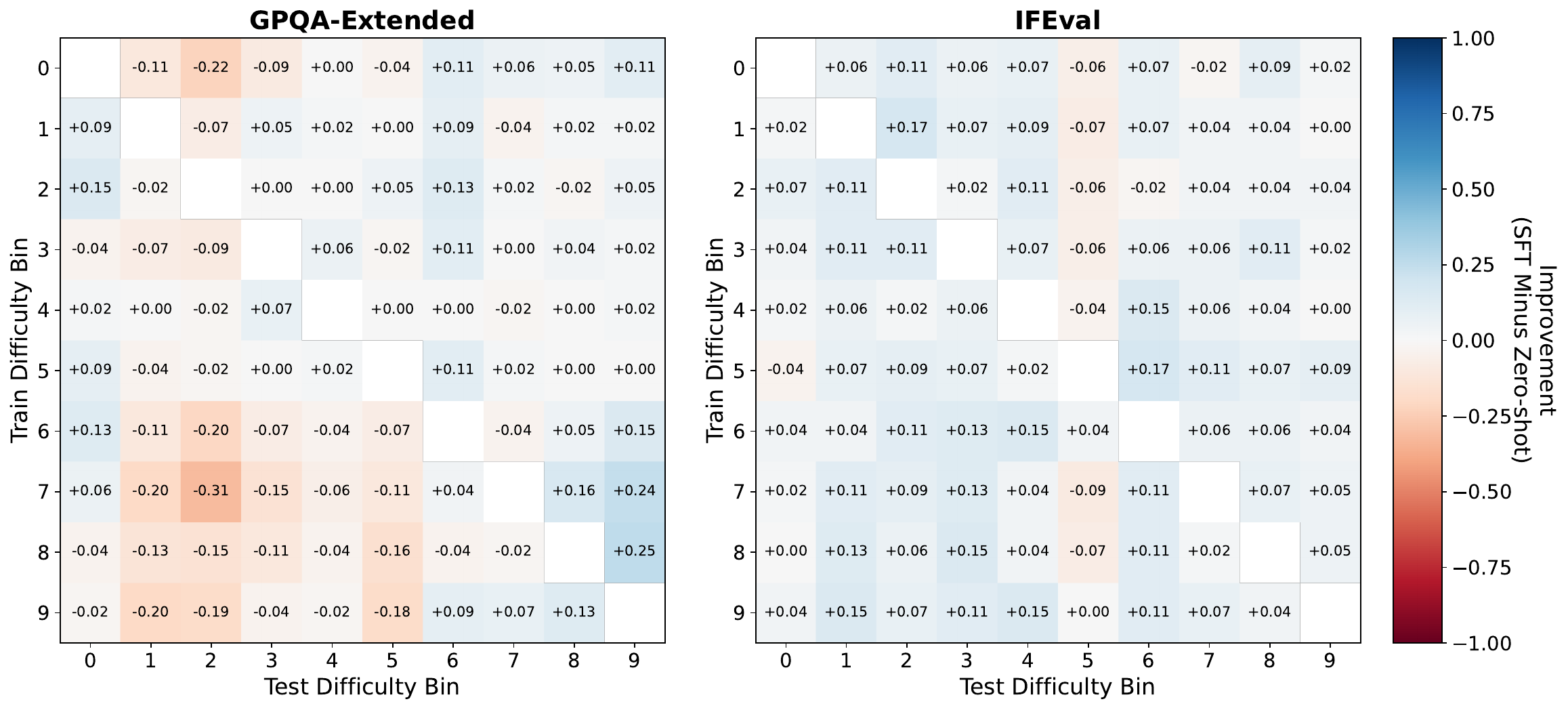}
    \caption{\textbf{Improvement analysis on IFEval and GPQA-Extended for Llama3.2 1B Instruct showing the difference between SFT and zero-shot performance.} Blue indicates positive improvements (SFT better than zero-shot), red indicates negative improvements (SFT worse than zero-shot).}
    \label{fig:llama1b_extra}
\end{figure*}

\section{Correlation}
\label{app:corr}
\begin{figure*}[htbp]
    \centering
    \includegraphics[width=\linewidth]{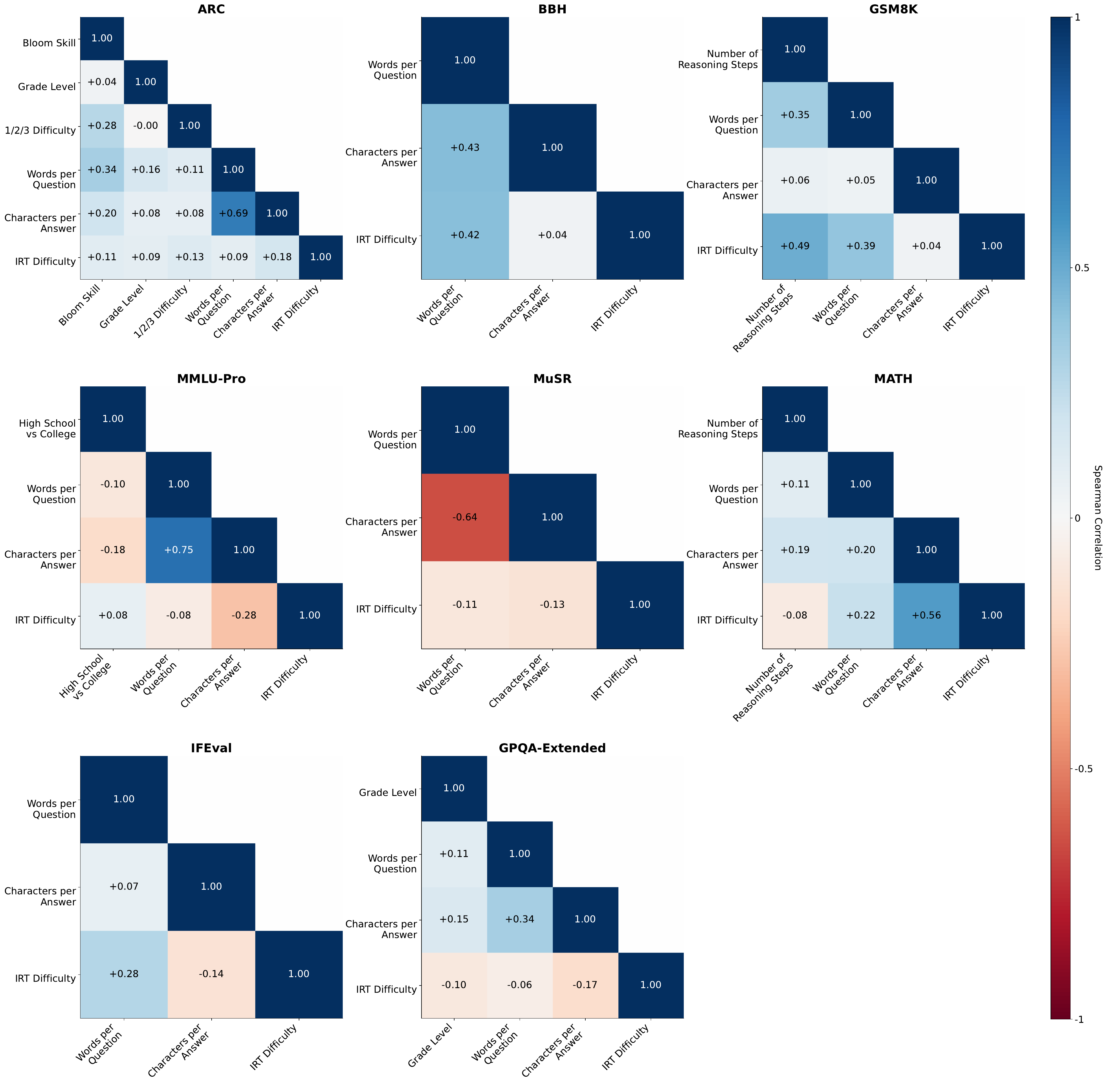}
    \caption{\textbf{Correlation heatmaps between IRT-based difficulty scores and human-defined metrics across all eight evaluation datasets.} Each heatmap shows Spearman rank correlation coefficients, with colors ranging from red (negative correlation) to blue (positive correlation). Metrics vary by dataset based on available annotations: ARC includes grade level and Bloom's taxonomy; GSM8K includes reasoning steps; MMLU-Pro includes high school vs. college labels. Across all datasets, most correlations remain weak ($|\rho| < 0.3$), with only GSM8k's reasoning steps ($\rho = 0.49$) and question length in BBH ($\rho = 0.42$) showing moderate positive correlations. Answer length consistently shows negative correlations, while expert-assigned difficulty ratings and educational levels show minimal alignment with model performance patterns.}
    \label{fig:correlations_between_hardness_measures}
\end{figure*}

In the main paper, we summarized the correlation between IRT-based difficulty scores and human-defined metrics (\S\ref{cross_diff:correlation}).
Here, we include the full set of correlation heatmaps across all eight evaluation datasets for completeness.
These extended results provide a more detailed view of how IRT-based difficulty correlates with various human annotations (e.g., grade level, reasoning steps, question and answer length).
As shown in Figure~\ref{fig:correlations_between_hardness_measures}, the overall correlations remain weak, consistent with the trends discussed in the main text.

\section{Examples}
\label{app:examples}
Table~\ref{tab:case_study1} shows some examples where the dataset's difficulty annotation disagrees with the IRT-based difficulty estimation, along with the models' performance on those instances.
Also, tables~\ref{tab:examples} and~\ref{tab:examples2} show additional examples from each IRT difficulty bin.
\input{tables/examples}

\input{tables/table2}

\clearpage
\section{Use of Large Language Models}
We used AI Assistants such as ChatGPT and Grammarly for spell-checking, fixing minor
grammatical mistakes, and polishing the writing. We also use GitHub CoPilot in VSCode to help write our codebase.

%% file: tables/examples.tex
\setlength\tabcolsep{3 pt} 
\newcommand{\sep}{-0.1cm} 
\newcommand{\sty}{\tt \scriptsize}
\renewcommand\labelitemi{--} 

\newcommand{\q}{{\greentext{Question:}}}
\newcommand{\opt}{\newline{\greentext{Options:}}}

\newcommand{\easy}{\textcolor{green!50!black}{\textit{(easy)}}}
\newcommand{\hard}{\textcolor{red!90!black}{\textit{(hard)}}}

\begin{table*}[th!]
\centering
\small
\resizebox{0.99\textwidth}{!}{
\begin{tabular}{p{0.4\textwidth} c p{0.3\textwidth} p{0.2\textwidth}}
\toprule
\multicolumn{1}{c}{\textbf{Question}}   & \multicolumn{1}{c}{\textbf{Answer}} & \multicolumn{1}{c}{\textbf{Difficulty ratings}}   & \multicolumn{1}{c}{\textbf{Model checklist}}\\
\midrule
\sty
\q{} A ball is rolling on the ground. A force pushes the ball in the same direction that it is moving. What happens to the ball? 
\opt{} 
\begin{itemize}[leftmargin=0.3cm, itemsep=\sep, topsep=0.1pt]
     \item A) It stops moving.
     \item B) It moves slower.
     \item C) It moves faster in the same direction it was moving.
     \item D) It keeps moving in the same speed and direction. 
\end{itemize}
 &  
 \sty C & 
 \vspace{-2mm}
 \sty
 \begin{itemize}[leftmargin=0.3cm, itemsep=\sep, topsep=0.1pt]
     \item ARC-Challenge: 3rd Grade \easy
     \item IRT: bin 9 \hard
 \end{itemize} & \sty 
 \vspace{-2mm}
 \begin{itemize}[leftmargin=0.3cm, itemsep=\sep, topsep=0.1pt]
    \item Llama3 70B: \xmark 
     \item Qwen2.5 72B: \xmark
     \item Mixtral 8x7B v0.1: \xmark
 \end{itemize}
 \\
\midrule
\sty
\q{} A major polling organization wants to predict the outcome of an upcoming national election (in terms of the proportion of voters who will vote for each candidate). They intend to use a $95\%$ confidence interval with margin of error of no more than $2.5\%$. What is the minimum sample size needed to accomplish this goal?
\opt{} 
\vspace{-\baselineskip}
\begin{multicols}{2}
\begin{itemize}[leftmargin=0.3cm, itemsep=\sep, topsep=0.1pt]
    \item A) 2048
    \item B) 1000
    \item C) 39
    \item D) 1536
    \item E) 2000
    \item F) 40
    \item G) 1024
    \item H) 1537
    \item I) 4096
    \item J) 500
\end{itemize}
\end{multicols}
 &  
 \sty H & 
 \vspace{-2mm}
 \sty
 \begin{itemize}[leftmargin=0.3cm, itemsep=\sep, topsep=0.1pt]
     \item MMLU-Pro: High School \easy
     \item IRT: bin 9 \hard
 \end{itemize} & \sty 
 \vspace{-2mm}
 \begin{itemize}[leftmargin=0.3cm, itemsep=\sep, topsep=0.1pt]
    \item Llama3 70B: \xmark 
     \item Qwen2.5 72B: \xmark
     \item Mixtral 8x7B v0.1: \xmark
 \end{itemize}
 \\
  \midrule
 \sty
\q{} A ligand binds to a deep beta-barrel cleft of an enzyme. Based on the 2.5 Angstrom resolution of the X-ray diffraction crystal structure, the amino acids in contact with the ligand at the active site seem to be H34, T48, S62, and G128. You would like to perform PCR mutagenesis of the enzyme's sequence to verify the mechanism of the ligand-receptor binding. Which of the point mutations below will most likely affect the ligand-receptor interaction?
\opt{} 
\vspace{-\baselineskip}
\begin{multicols}{2}
\begin{itemize}[leftmargin=0.3cm, itemsep=\sep, topsep=0.1pt]
     \item A) 98A $\rightarrow$ G
     \item B) 128G $\rightarrow$ C
     \item C) 142A $\rightarrow$ G
     \item D) 186T $\rightarrow$ C
\end{itemize}
\end{multicols}
 &  
 \sty C & 
 \vspace{-2mm}
 \sty
 \begin{itemize}[leftmargin=0.3cm, itemsep=\sep, topsep=0.1pt]
     \item GPQA-Extended: Easy Undergrad \easy
     \item IRT: bin 9 \hard
 \end{itemize} & \sty 
 \vspace{-2mm}
 \begin{itemize}[leftmargin=0.3cm, itemsep=\sep, topsep=0.1pt]
    \item Llama3 70B: \xmark 
     \item Qwen2.5 72B: \xmark
     \item Mixtral 8x7B v0.1: \xmark
 \end{itemize}
 \\
 \midrule
\sty
\q{} A potential negative impact of building a dam on a river is that the dam
\opt{} 
\begin{itemize}[leftmargin=0.3cm, itemsep=\sep, topsep=0.1pt]
     \item A) prevents sediment from flowing downstream.
     \item B) increases the amount of water available to farms.
     \item C) prevents seasonal downstream flooding.
     \item D) increases the rate of water loss from a lake.
\end{itemize}
 &  
 \sty A & 
 \vspace{-2mm}
 \sty
 \begin{itemize}[leftmargin=0.3cm, itemsep=\sep, topsep=0.1pt]
     \item ARC-Challenge: 8th Grade \hard
     \item IRT: bin 0 \easy
 \end{itemize} & \sty 
 \vspace{-2mm}
 \begin{itemize}[leftmargin=0.3cm, itemsep=\sep, topsep=0.1pt]
    \item Llama3 70B: \cmark 
     \item Qwen2.5 72B: \cmark
     \item Mixtral 8x7B v0.1: \cmark
 \end{itemize}
 \\
 \midrule
 \sty
\q{} A new enzyme is found in a transgenic mice that participates in synthesis of an unknown product using two reactants. When using radiolabeled compounds to study the enzyme, it is found that the enzyme catalyzes a process that switches a nitrogen group on one reactant to the other reactant. Which of the following categories would this new enzyme fall under?
\opt{} 
\vspace{-\baselineskip}
\begin{multicols}{2}
\begin{itemize}[leftmargin=0.3cm, itemsep=\sep, topsep=0.1pt]
     \item A) Ligase
     \item B) Hydrolase
     \item C) Transferase
     \item D) Synthetase
     \item E) Phosphatase
     \item F) Lyase
     \item G) Oxidoreductase
     \item H) Decarboxylase
     \item I) Kinase
     \item J) Isomerase
\end{itemize}
\end{multicols}
 &  
 \sty C & 
 \vspace{-2mm}
 \sty
 \begin{itemize}[leftmargin=0.3cm, itemsep=\sep, topsep=0.1pt]
     \item MMLU-Pro: College \hard
     \item IRT: bin 0 \easy
 \end{itemize} & \sty 
 \vspace{-2mm}
 \begin{itemize}[leftmargin=0.3cm, itemsep=\sep, topsep=0.1pt]
    \item Llama3 70B: \cmark 
     \item Qwen2.5 72B: \cmark
     \item Mixtral 8x7B v0.1: \cmark
 \end{itemize}
 \\
 \midrule
 \sty
\q{} What is a distinctive feature of thin film deposition using High Power Impulse Magnetron Sputtering (HiPIMS)?
\opt{} 
\begin{itemize}[leftmargin=0.3cm, itemsep=\sep, topsep=0.1pt]
     \item A) HiPIMS primarily relies on chemical vapor deposition (CVD) mechanisms for film formation.
     \item B) HiPIMS generates short, intense plasma pulses to enhance ionization and film growth.
     \item C) HiPIMS utilizes continuous DC power for a consistent deposition rate.
     \item D) HiPIMS operates at lower vacuum pressures compared to traditional sputtering techniques.
\end{itemize}
 &  
 \sty B & 
 \vspace{-2mm}
 \sty
 \begin{itemize}[leftmargin=0.3cm, itemsep=\sep, topsep=0.1pt]
     \item GPQA-Extended: Post Graduate \hard
     \item IRT: bin 0 \easy
 \end{itemize} & \sty 
 \vspace{-2mm}
 \begin{itemize}[leftmargin=0.3cm, itemsep=\sep, topsep=0.1pt]
    \item Llama3 70B: \cmark 
     \item Qwen2.5 72B: \cmark
     \item Mixtral 8x7B v0.1: \cmark
 \end{itemize}
 \\
 \bottomrule
\end{tabular}
}
\caption{Examples of Disagreement Between Dataset Difficulty Label and IRT Difficulty}
\label{tab:case_study1}
\end{table*}

%% file: tables/table2.tex
\setlength\tabcolsep{3 pt} 
\renewcommand\labelitemi{--} 

\begin{table*}[th!]
\centering
\small
\resizebox{0.99\textwidth}{!}{
\begin{tabular}{p{0.5\textwidth} p{0.3\textwidth} c p{0.1\textwidth}}
\toprule
\multicolumn{1}{c}{\textbf{Question}}   & \multicolumn{1}{c}{\textbf{Answer}} & \multicolumn{1}{c}{\textbf{Dataset}} & \multicolumn{1}{c}{\textbf{Difficulty ratings}}\\
\midrule
\sty
\q{} ( not True ) and True and False is
 &  
 \sty False 
  &  
 \sty BBH
 & 
 \vspace{-2mm}
 \sty IRT: bin 0 \\
\midrule
\sty
\q{} There are 220 castles in Scotland. 40 percent of them are ruins, and half of the ruined castles are unmanned. How many unmanned ruined castles are there in Scotland?
 &  
 \sty  

Number of ruined castles $=$  
\[
0.40 \times 220 = 88
\]
Half of the ruined castles are unmanned. Number of unmanned ruined castles $=$
\[
\frac{88}{2} = 44
\]
 & 
 \sty GSM8K
 & 
 \sty IRT: bin 1 
 \\
  \midrule
 \sty
\q{} Which is most responsible for recycling dead plants and animals in an ecosystem?
\opt{} 
\begin{itemize}[leftmargin=0.3cm, itemsep=\sep, topsep=0.1pt]
     \item A) water
     \item B) bacteria
     \item C) fish
     \item D) insects
\end{itemize}
 &  
 \sty B
 &  
 \sty ARC
 & 
 \sty IRT: bin 2
 \\
 \midrule
\sty
\q{} Here is the summary of a research paper on the effect of VHF radio waves on a certain type of bacteria: "Our results show that VHF radio waves have no effect on the growth of bacteria." Can you help me rewrite this summary in a more formal way, using APA format? Do not use words "ours" or "have".
 &  
 \sty Here's a rewritten summary in a more formal tone, following APA format:

"The results of this study indicate that exposure to VHF radio waves does not affect the growth of bacteria."

Alternatively, you could also phrase it as:

"No significant effect on bacterial growth was observed following exposure to VHF radio waves."

Both of these rewritten summaries maintain a formal tone and adhere to APA format guidelines.
 &   
 \sty IFEval
 & 
 \sty IRT: bin 3
 \\
 \midrule
 \sty
\q{} In the vibrant realm of music, the responsibility of delegating roles fell upon my shoulders. My ensemble consisted of three distinct personalities - Alice, Michael, and Henry. Each brought their own unique flair and abilities to the table, making the task of assigning roles a captivating puzzle. The primary roles to be allocated were that of an instrumentalist and a singer. The unique attributes of each band member meant that the decision required careful consideration and balance.

Alice was a complex character. Though she was not one to hold eye contact with Michael during performances, she carried with her a passion for guitar that few could rival. Daily, she spent at least an hour dedicated solely to practicing her guitar skills. Her devotion was evident from the fact that she had been part of her school band as a guitarist. However, while Alice was eager to contribute, her voice would often falter during moments of singing. She'd lose her voice within a short span of singing, probably due to lack of formal vocal training.

Henry, on the other hand, was quite a handful. He often interrupted Alice, largely ignoring her efforts to contribute to their shared passion. Henry might have been skilled, but he showed little interest in bettering his abilities on the bass guitar. His lack of practice was evident during the last gig when he was unable to keep up with the tempo, stopping midway through the performance. Perhaps his singing suffered as a result. Even though Henry had never taken singing lessons, his voice tended to crack often when he sang, indicating a lack of vocal control.

Michael, whilst being a firecracker in band meetings, had his shortcomings as well. He had a knack for creativity which Henry often unjustly dismissed. Surprisingly, Michael failed to deliver during the last performance where he was unable to hit the high notes during his solo. He also visibly struggled to maintain the rhythm of his drum solos and often forgot his lines mid-song. Michael's odd trait was his tendency to forget his lines when Alice was around, an unfortunate condition considering they were bandmates. This became painfully clear when he failed to remember a simple drum solo he was assigned.

The unraveling of the intricacies of each band member, their dynamics with one another, their strengths and weaknesses were all fascinating to say the least. Assigning them to either singing or instrumental duties was a task fraught with careful analysis. But through it all, one thing was for certain - the spark of artistry was present in each one of them. The contours of the task lay ahead with an air of anticipation, offering an unique challenge that I was ready to surmount. The music played on, and so did the story of Alice, Michael and Henry. 
Given the story, how would you uniquely allocate each person to make sure both tasks are accomplished efficiently?
 &  
 \sty Instrumentalist: Alice, Singer: Henry and Michael 
 &  
 \sty MuSR
 & 
 \sty IRT: bin 4
 \\
 \bottomrule
\end{tabular}
}
\caption{Examples of Questions from IRT difficulty bins 0–4}
\label{tab:examples}
\end{table*}

\begin{table*}[th!]
\centering
\small
\resizebox{0.99\textwidth}{!}{
\begin{tabular}{p{0.5\textwidth} p{0.3\textwidth} c p{0.1\textwidth}}
\toprule
\multicolumn{1}{c}{\textbf{Question}}   & \multicolumn{1}{c}{\textbf{Answer}} & \multicolumn{1}{c}{\textbf{Dataset}} & \multicolumn{1}{c}{\textbf{Difficulty ratings}}\\
\midrule
\sty
\q{} Solve for $x$: $$\dfrac{66-2^x}{2^x+3}=\dfrac{4-2^x}{2^{x+1}+6}$$
 &  
 \sty First, we recognize that $2^{x+1}+6=2(2^x+3)$: $$\dfrac{2(66-2^x)}{2(2^x+3)}=\dfrac{4-2^x}{2(2^x+3)}$$Then, we expand and collect like terms: $$\dfrac{128-2^x}{2(2^x+3)} = 0$$This equation can only be true when $2^x = 128$, which indicates that $x = \boxed{7}$.
 &   
 \sty MATH
 & 
 \sty IRT: bin 5
 \\
 \midrule
\sty
\q{} 	
A researcher interested in examining the potential impact of parent alcoholism on child and family development recruits 12-year-olds (n = 100), 13-year-olds (n = 100), and 14-year-olds (n = 100)—half of whom have an alcoholic parent and half of whom do not—into a multiple-year longitudinal study assessing various outcomes. This study is best characterized as:
\opt{} 
\vspace{-\baselineskip}
\begin{multicols}{2}
\begin{itemize}[leftmargin=0.3cm, itemsep=\sep, topsep=0.1pt]
     \item A) A cross-sectional design
     \item B) A correlational study
     \item C) A pretest-posttest design
     \item D) A cross-sequential cohort design
     \item E) A quasi-experiment
     \item F) A natural experiment
     \item G) A cross-sectional cohort design
     \item H) A true experiment
     \item I) A case-control study
     \item J) A longitudinal cohort design
\end{itemize}
\end{multicols}
 &  
 \sty D
 &   
 \sty MMLU-Pro
 & 
 \sty IRT: bin 6
 \\
 \midrule
\sty
\q{} methyl (E)-but-2-enoate is treated with quinuclidine and acetone, forming product 1.

1 is treated with excess methylmagnesium bromide, forming product 2.

how many chemically distinct non-exchanging hydrogen signals will there be in the 1H nmr spectrum of product 2? (There may be signals that practically would have very close chemical shifts, but the answer should be the number that are in principle distinguishable.)
\opt{} 
\begin{itemize}[leftmargin=0.3cm, itemsep=\sep, topsep=0.1pt]
     \item A) 8
     \item B) 6
     \item C) 3
     \item D) 4
\end{itemize}
 &  
 \sty D
 &   
 \sty GPQA-Extended
 & 
 \sty IRT: bin 7
 \\
 \midrule
\sty
\q{} Two jokers are added to a $52$ card deck and the entire stack of $54$ cards is shuffled randomly. What is the expected number of cards that will be strictly between the two jokers?
 &  
 \sty Each card has an equal likelihood of being either on top of the jokers, in between them, or below the jokers. Thus, on average, $1/3$ of them will land between the two jokers. Multiplying this by the 52 yields our answer of $\boxed{\frac{52}{3}}$.
 &   
 \sty MATH
 & 
 \sty IRT: bin 8
 \\
 \midrule
\sty
\q{} 	
Alice, Bob, Claire, Dave, Eve, Fred, and Gertrude are dancers at a square dance. At the start of a song, they each have a partner: Alice is dancing with Ophelia, Bob is dancing with Melissa, Claire is dancing with Jamie, Dave is dancing with Sam, Eve is dancing with Patrick, Fred is dancing with Rodrigo, and Gertrude is dancing with Karl.
Throughout the song, the dancers often trade partners. First, Dave and Claire switch partners. Then, Alice and Eve switch partners. Then, Eve and Bob switch partners. Then, Claire and Bob switch partners. Then, Fred and Eve switch partners. Then, Gertrude and Dave switch partners. Finally, Dave and Alice switch partners. At the end of the dance, Fred is dancing with
\opt{} 
\begin{itemize}[leftmargin=0.3cm, itemsep=\sep, topsep=0.1pt]
     \item A) Ophelia
     \item B) Melissa
     \item C) Jamie
     \item D) Sam
     \item E) Patrick
     \item F) Rodrigo
     \item G) Karl
\end{itemize}
 &  
 \sty B
 &   
 \sty BBH
 & 
 \sty IRT: bin 9
 \\
  \bottomrule
\end{tabular}
}
\caption{Examples of Questions from IRT difficulty bins 5-9}
\label{tab:examples2}
\end{table*}